\documentclass[10pt,twocolumn,letterpaper]{article}

\usepackage[pagenumbers]{cvpr} %

\definecolor{cvprblue}{rgb}{0.21,0.49,0.74}
\usepackage[pagebackref,breaklinks,colorlinks,allcolors=cvprblue]{hyperref}

\usepackage{times}
\usepackage{epsfig}
\usepackage{graphicx}
\usepackage{amsmath}
\usepackage{amssymb}
\usepackage{booktabs}
\usepackage{bm}
\usepackage{comment}
\usepackage{mathtools}
\usepackage{xfrac}
\usepackage{multirow}
\usepackage{nicefrac}
\usepackage{stmaryrd}
\usepackage{lipsum}  
\usepackage{csquotes}  
\usepackage{multicol}
\usepackage{lipsum}

\colorlet{best3}{red!20!white}
\colorlet{best2}{yellow!50!white}
\colorlet{best1}{green!30!white}

\def \bestA {\colorbox{best1}}
\def \bestB {\colorbox{best2}}
\def \bestC {\colorbox{best3}}

\definecolor{high}{HTML}{76f013}  %
\definecolor{low}{HTML}{ec462e}  %

\newcommand{\Acronym}{MVGD\xspace}%

\def\paperID{1440} %
\def\confName{CVPR}
\def\confYear{2025}

\title{
Zero-Shot Novel View and Depth Synthesis 
with Multi-View Geometric Diffusion
}

\author{
Vitor Guizilini$^1$ \quad\quad 
Muhammad Zubair Irshad$^1$ \quad\quad 
Dian Chen$^1$ \vspace{1mm} \\ 
Greg Shakhnarovich$^2$ \quad\quad 
Rares Ambrus$^1$ 
\vspace{3mm} \\
Toyota Research Institute (TRI)$^1$  \quad 
Toyota Technological Institute at Chicago (TTIC)$^2$ 
}

\begin{document}
\twocolumn[{%
\renewcommand\twocolumn[1][]{#1}%
\maketitle
\vspace{-14mm}
\begin{center}
    \centering
    \includegraphics[width=0.24\textwidth,height=3.5cm]{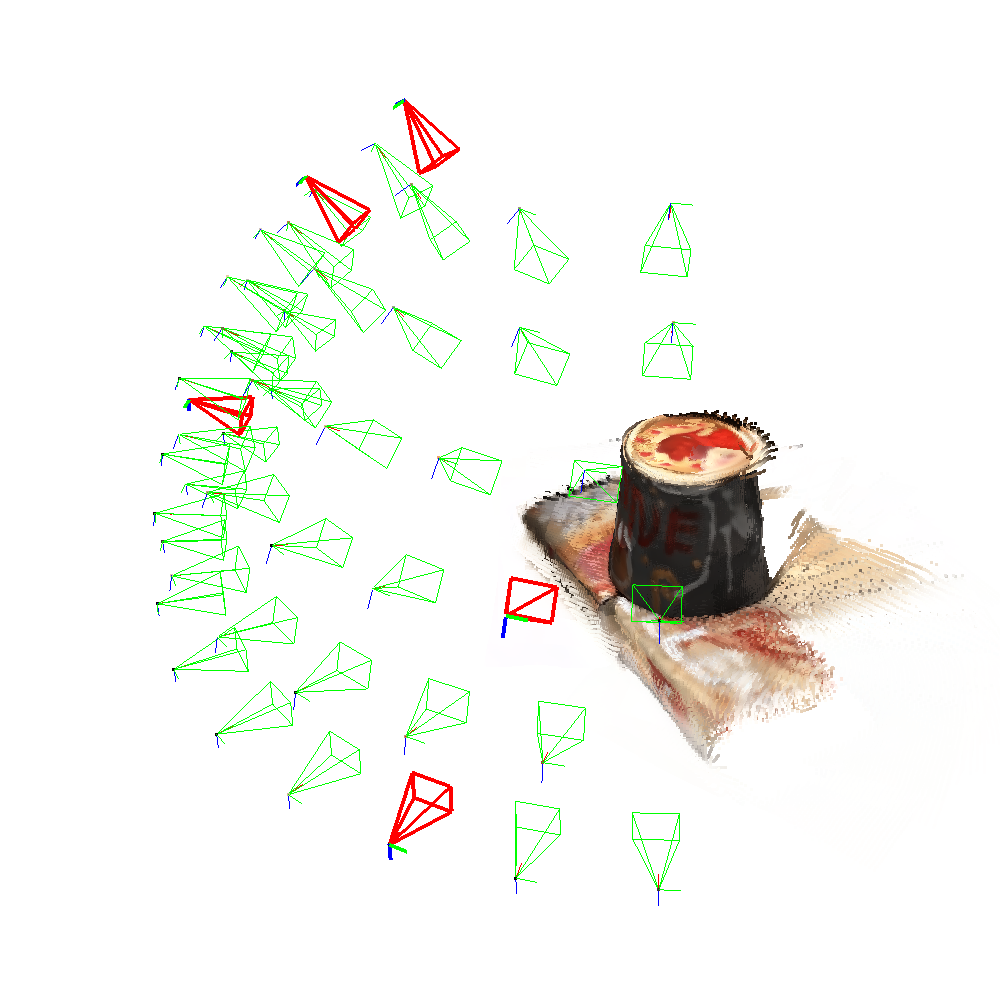}
    \includegraphics[width=0.24\textwidth,height=3.5cm]{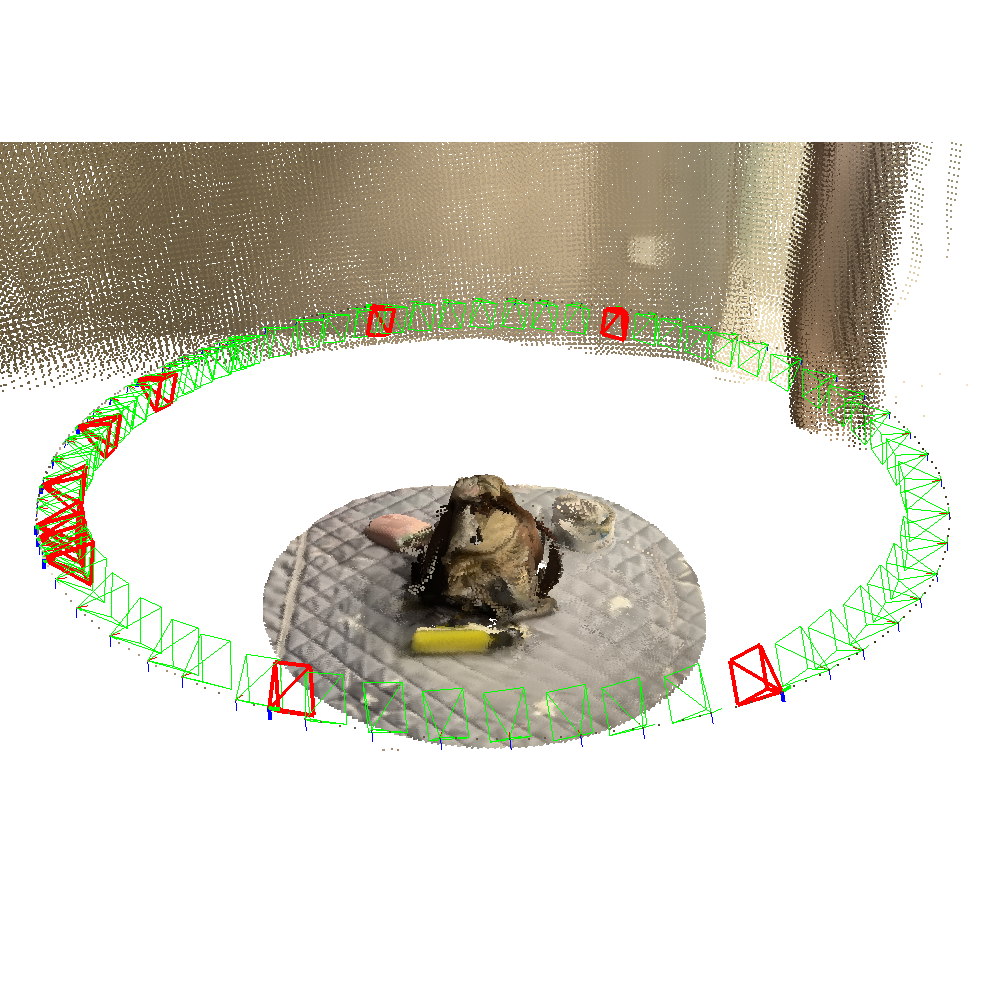}
     \includegraphics[width=0.24\textwidth,height=3.5cm]{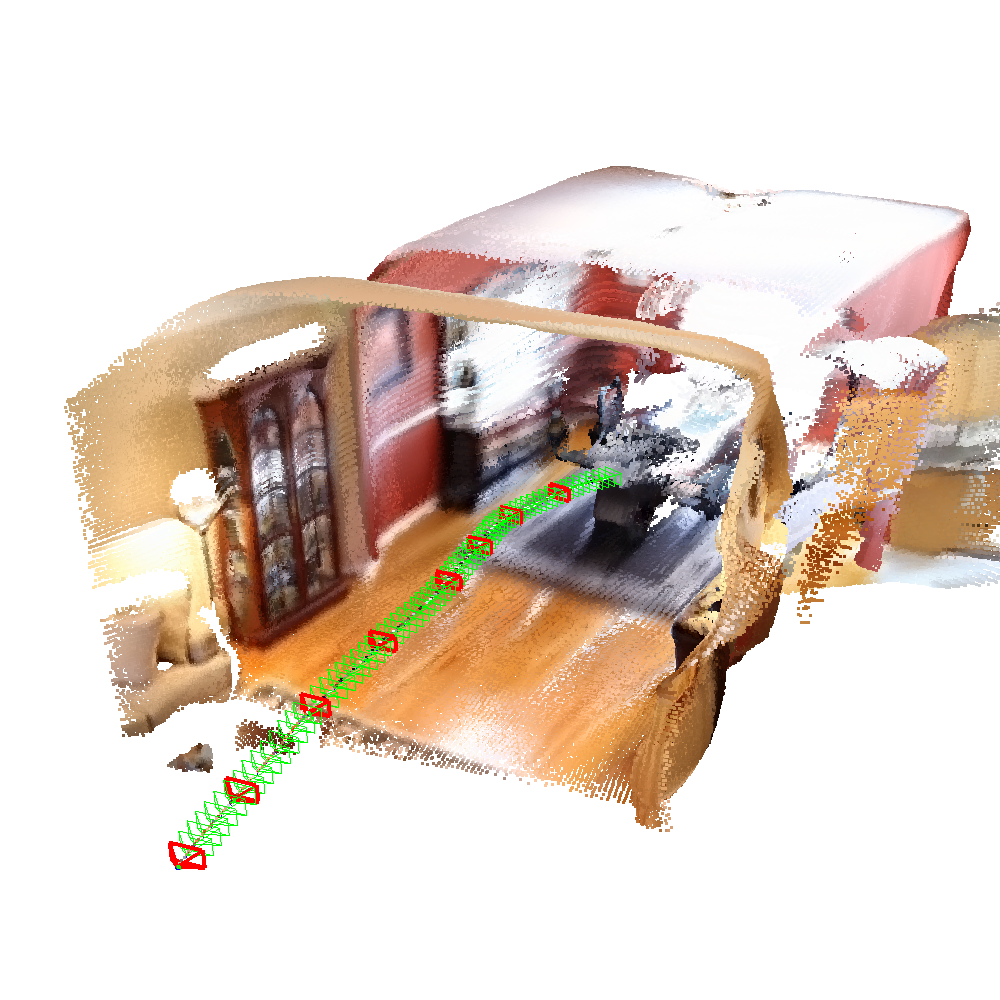}
    \includegraphics[width=0.24\textwidth,height=3.5cm]{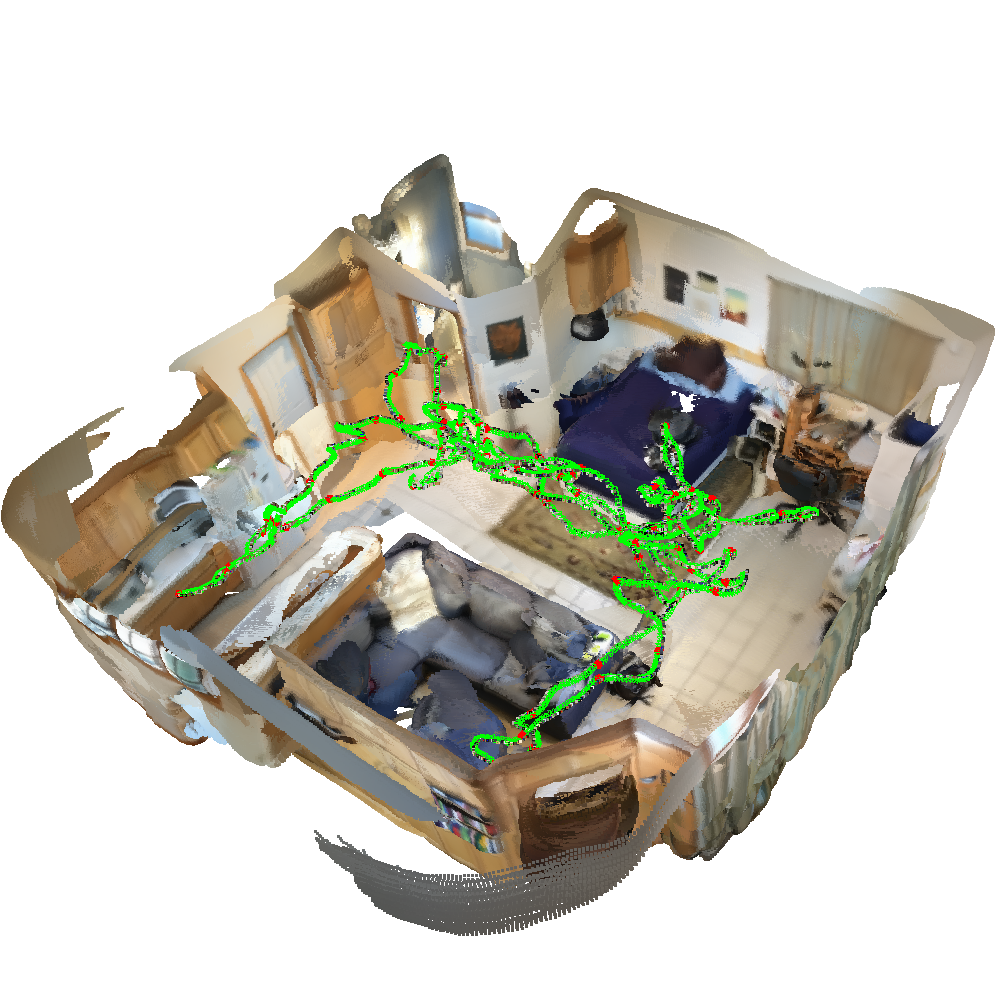}
    \captionsetup{width=\linewidth}
    \vspace{-6mm}
    \captionof{figure}{
        \textbf{\Acronym is a state-of-the-art method that generates 
        images and scale-consistent depth maps from novel viewpoints} given an arbitrary number of posed input views. In the above, red cameras are used as conditioning to directly generate RGB-D predictions from green cameras. To highlight the multi-view consistency of our method, predicted colored pointclouds from all novel viewpoints are stacked together for visualization without any post-processing. More examples and videos can be found in \href{https://mvgd.github.io/}{https://mvgd.github.io/}
    }
   \label{fig:teaser}
\end{center}%
\vspace{-1mm}
}]

\begin{abstract}
Current methods for 3D scene reconstruction from sparse posed images employ intermediate 3D representations such as neural fields, voxel grids, or 3D Gaussians, to achieve multi-view consistent scene appearance and geometry.
In this paper we introduce \Acronym, a diffusion-based architecture capable of direct pixel-level generation of images and depth maps from novel viewpoints, given an arbitrary number of input views. %
Our method uses raymap conditioning to both augment visual features with spatial information from different viewpoints, as well as to guide the generation of images and depth maps from novel views.
A key aspect of our approach is the multi-task generation of images and depth maps, using learnable task embeddings to guide the diffusion process towards specific modalities.
We train this 
model 
on a collection of more than 60 million multi-view samples from publicly available datasets, and propose techniques to enable efficient and consistent learning in such diverse conditions. 
We also propose a novel strategy that enables the efficient training of larger models by incrementally fine-tuning smaller ones, with promising scaling behavior.
Through extensive experiments, we report state-of-the-art results in multiple novel view synthesis benchmarks, as well as multi-view stereo and video depth estimation.

\end{abstract}
\vspace{-5mm}
\section{Introduction}

We investigate the task of generalizable novel view synthesis and depth estimation from sparse posed images. 
This task has received much attention 
with the advent of Neural Radiance Fields (NeRFs)~\cite{mildenhall2020nerf}, 3D Gaussian Splatting (3DGS)~\cite{kerbl3Dgaussians} and other neural methods~\cite{sitzmann2021lfns,srt22,tri-define} for constructing a parameterized 3D representation of the scene observed from different viewpoints. Novel views, and by extension depth maps observed from novel viewpoints, can be generated via volumetric rendering or rasterization from the 3D model, once it is constructed from available input views.
Subsequent work introduced improvements including generalization to novel scenes~\cite{pixelnerf,charatan23pixelsplat,chen2024mvsplat}, better sample-efficiency~\cite{dsnerf,nerfingmvs,roessle2022depthpriorsnerf}, reduced memory cost~\cite{tri-delira}, and rendering speedups~\cite{tri-delira,sitzmann2021lfns}.  Despite these advances, the ability to generalize to novel scenes remains limited, especially with increasing numbers of input views~\cite{charatan23pixelsplat,chen2024mvsplat}. %

We pursue a compelling alternative to creating a parameterized 3D representation: \textbf{the direct rendering of novel views and depth maps as a conditional generative task}. 
Our work is inspired by light and depth field networks~\cite{sitzmann2021lfns,srt22, tri-define, tri-delira} and recent developments on multi-view diffusion~\cite{wu2023reconfusion,hu2024mvdfusion,xu2023dmv3d,gao2024cat3d}. Diffusion-based methods for novel view synthesis have so far been limited to simpler settings~\cite{liu2023zero1to3,liu2023syncdreamer} or as a source of 2D priors for downstream 3D reconstruction~\cite{wu2023reconfusion,gao2024cat3d}. This is largely due to a key challenge: the difficulty, without an intermediate 3D scene representation, to impose multi-view consistency when generating novel views.
Our proposed solution to this problem is to train a single diffusion model for the joint task of novel \emph{view} and \emph{depth} synthesis, using direct conditioning on the input views and known input camera geometry via \emph{raymap conditioning}. We show that this leads to the generation of scale-aware and multi-view consistent predictions without any intermediate 3D representation or special mechanisms. %

To train such a model that learns to implicitly represent multi-view consistent 3D scenes from sparse observations,
we curated more than 60M multi-view samples~(see Tab.~\ref{tab:datasets}) from publicly available datasets, including real-world and synthetic images from a widely diverse range of driving, indoors, and robotics scenarios. 
To properly leverage such diverse data, we introduce novel techniques for effective training in heterogeneous conditions.
First, to handle possibly sparse real-world depth, we use an efficient Transformer architecture~\cite{rin} to perform pixel-level diffusion, thus removing the need for dedicated auto-encoders~\cite{grin}.
Second, we use learnable task embeddings to determine which output should be generated from the same conditioned latent tokens, enabling unified training on datasets with or without depth labels. 
Lastly, to handle non-metric datasets, we propose a novel scene scale normalization (SSN) procedure, to generate scale-aware and multi-view consistent depth maps. These strategies enable the effective scaling of our method, resulting in robust generalization across diverse scenarios. %

Our proposed \emph{Multi-View Geometric Diffusion (\Acronym)} framework outperforms existing baselines on several in-domain and zero-shot benchmarks, establishing a new state-of-the-art in novel view synthesis with varying amounts of conditioning views (2-9). 
To further showcase our implicit geometric reasoning, we also evaluate \Acronym for the tasks of stereo and video depth estimation, reporting state of the art results on the ScanNet benchmark.
We show that \Acronym can process arbitrarily large numbers of conditioning views ($100+$), and propose an incremental conditioning strategy designed to further improve multi-view consistency over long sequences. Finally, we report promising scaling behavior in which increasing model complexity improves results, and present a strategy for incremental fine-tuning from smaller models that cuts down on training time by up to $70\%$.
In summary, our contributions are as follows:
\begin{itemize}
    \item A \textbf{novel multi-task diffusion architecture (MVGD)} capable of multi-view consistent image and depth synthesis from an arbitrary number of posed images (up to thousands) without an intermediate 3D representation.
    \item A set of techniques designed to \textbf{enable training on a large-scale, heterogeneous dataset with over 60M multi-view samples} from diverse visual domains, different levels of 3D supervision, varying numbers of views and cameras, and a broad range of scene scales.
    \item A strategy for the \textbf{efficient training of larger \Acronym models} by incrementally fine-tuning smaller ones, with promising scaling behavior in terms of performance.  
\end{itemize}

\noindent The combined effect of these contributions allows us to establish a new state of the art in generalizable novel view synthesis, as well as multi-view stereo and video depth estimation, on a number of well established benchmarks.

\begin{table}[t!]
\scriptsize
\renewcommand{\arraystretch}{0.85}
\centering
\setlength{\tabcolsep}{0.2em}
\begin{tabular}{l|c|c|c|c|c|r|r}
\toprule
Dataset & Syn. & Dyn. & Met. & I/O & V/M & \# Sequences & \# Samples \\
\midrule
ArgoVerse2~\cite{Argoverse2}
&               & \checkmark    & \checkmark & O  & VM &    $1043$ &  $3,909,297$ \\
BlendedMVG~\cite{yao2020blendedmvs}
&               &               &            & O  &  M &     $502$ &    $115,142$ \\
CO3Dv2~\cite{reizenstein21co3d}
&               &               &            & I  &  V &  $25,243$ &  $5,088,873$ \\
DROID~\cite{khazatsky2024droid}
&               &               & \checkmark & I  &  M &  $76,792$ &  $7,340,712$ \\
LSD~\cite{tri-zerodepth}
&               & \checkmark    & \checkmark & O  & VM &  $17,647$ &  $1,057,920$ \\
LyftL5~\cite{lyftl5}
&               & \checkmark    & \checkmark & O  & VM &     $394$ &    $347,508$ \\
MVImgNet~\cite{yu2023mvimgnet}
&               &               &            & I  &  V & $194,368$ &  $5,768,120$ \\
NuScenes~\cite{caesar2020nuscenes}
&               & \checkmark    & \checkmark & O  & VM &     $850$ &    $204,894$ \\
Taskonomy~\cite{zamir2018taskonomy}
&               &               & \checkmark & O  &  M &     $533$ &  $4,584,462$ \\
HM3D~\cite{ramakrishnan2021hm3d}
&               &               & \checkmark & O  &  M &     $900$ &  $9,531,876$ \\
Hypersim~\cite{roberts:2021}
&               &               & \checkmark & O  &  M &     $457$ &     $74,619$ \\
P. Domain~\cite{guda,draft}
& \checkmark    & \checkmark    & \checkmark & IO & VM &   $5,449$ &    $555,000$ \\
RealEstate10k~\cite{zhou2018stereo}
&               &               &   ---      & O  &  V &  $74,532$ & $10,115,793$ \\
RTMV~\cite{tremblay2022rtmv}
& \checkmark    &               &            & I  &  M &   $1,909$ &    $286,350$ \\
ScanNet~\cite{dai2017scannet}
&               &               & \checkmark & O  &  V &   $1,513$ &  $2,477,378$ \\
TartanAir~\cite{wang2020tartanair}
& \checkmark    & \checkmark    & \checkmark & O  & VM &     $369$ &    $613,274$ \\
VKITTI2~\cite{cabon2020vkitti2}
& \checkmark    & \checkmark    & \checkmark & O  & VM &      $45$ &     $38,268$ \\
Waymo~\cite{waymo}
& \checkmark    &               & \checkmark & O  & VM &   $1,000$ &    $990,340$ \\
WildRGBD~\cite{xia2024rgbd}
&               &               &            & I  &  V &  $23,049$ &  $8,026,495$ \\
\midrule
\multicolumn{6}{l|}{\textbf{Total}}
& $\bm{426,595}$ & $\bm{61,126,321}$ \\
\bottomrule
\end{tabular}
\vspace{-2mm}
\caption{
\textbf{Datasets used to train \Acronym.} \emph{Syn} indicates if the dataset is synthetic, \emph{Dyn.} if it contains dynamic objects, \emph{Met.} if depth labels are metric, \emph{I/O} if cameras are looking inwards and/or outwards, and \emph{M/V} if it contains multiple cameras and/or videos.%
}
\label{tab:datasets}
\vspace{-6mm}
\end{table}

\section{Related Work}

\looseness=-1
\noindent\textbf{Generalizable Radiance Fields.} Traditionally, radiance field representations~\cite{muller2022instant,mildenhall2020nerf,kerbl3Dgaussians,irshad2024nerfmae,barron2021mipnerf} are trained at a per-scene basis, incurring expensive optimization. Generalizable formulations~\cite{pixelnerf,irshad2023neo360,charatan23pixelsplat} tackle the problem of few-shot view-synthesis, hence reducing their dependency on dense captures. These methods~\cite{pixelnerf, irshad2023neo360, srt22, trevithick2021grf, srt22, trevithick2021grf} utilize local features as conditioners in a feed-forward neural network for effective few-view synthesis. By leveraging a vast collection of multi-view posed data from various scenes, they can learn transferable priors. However, performance declines when the input views fall outside the distribution. More recently, this idea has been extended to 3D Gaussian Splatting~\cite{charatan23pixelsplat, chen2024mvsplat} for real-time few-shot rendering. Alternatively, other methods rely on regularizing the scene's geometry ~\cite{Niemeyer2021Regnerf, somraj2023simplenerf, wynn-2023-diffusionerf}, ensuring semantic consistency~\cite{dietnerf}, or utilizing sparse depth-supervision~\cite{dsnerf, roessle2022depthpriorsnerf, wang2023sparsenerf} to improve rendering quality given few views. 

\begin{figure*}[t!]
\vspace{-5mm}
\begin{center}
    \centering
\includegraphics[width=0.85\textwidth]{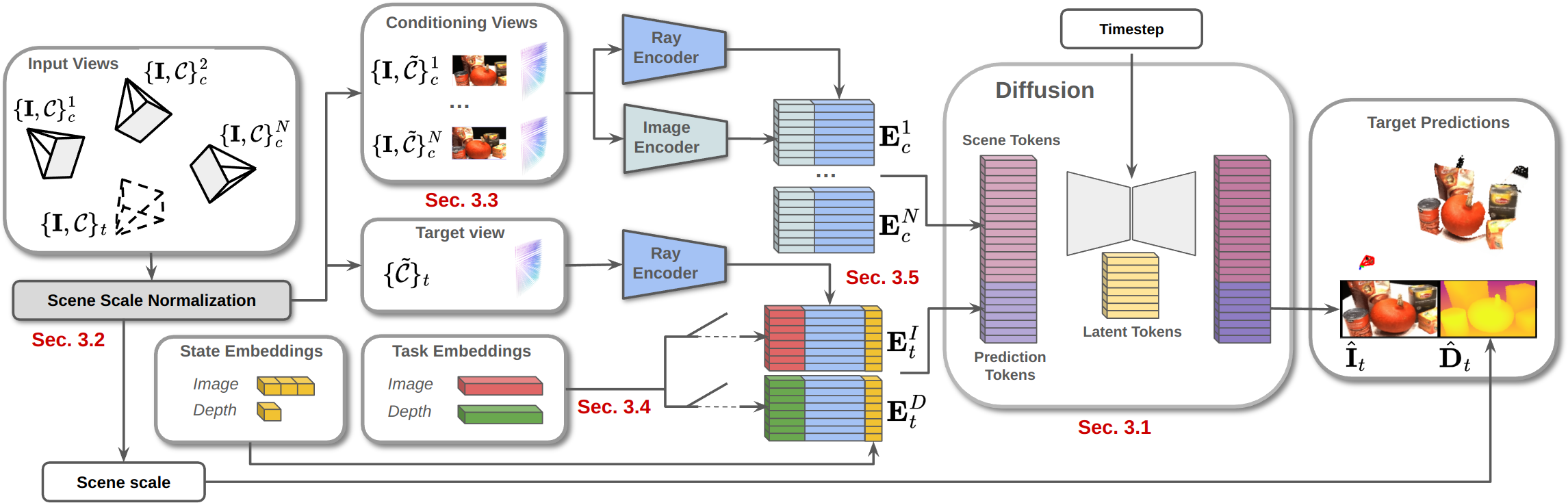}
\\
\vspace{-2mm}
\caption{
    \textbf{Diagram of our proposed Multi-View Geometric Diffusion (\Acronym) framework, at inference time.} %
    $N$ input images $\mathbf{I}_c^n$ with cameras $\mathcal{C}_c^n$ are used for scene conditioning, and a different camera $\mathcal{C}_t$ is selected for novel view and depth synthesis. %
}
\label{fig:diagram}
\vspace{-8mm}
\end{center}
\end{figure*}

\noindent\textbf{Generative Models for Novel View Synthesis.} Generating novel views outside the distribution of inputs views requires extrapolating unseen parts of the scene. Existing works have utilized Generative Adversarial Networks (GANs)~\cite{Chan2021, Niemeyer2020GIRAFFE, Schwarz2022NEURIPS, devries2021unconstrained, chanmonteiro2020pi-GAN} for conditional or unconditional scene generation. Owing to the impressive success of Diffusion models~\cite{ho2020denoising} for text-to-image generation~\cite{ramesh2021zero, rombach2022high}, recent works have utilized Diffusion Models~\cite{chan2023genvs,gu2023nerfdiff,karnewar2023holodiffusion,liu2023zero1to3, liu2023syncdreamer, tewari2023forwarddiffusion, yoo2023dreamsparse} for the task of novel view synthesis. Existing methods~\cite{liu2023zero1to3, watson2022novelviewsynthesisdiffusion} train the conditional diffusion model with relative poses as inputs and show great generalization on object-centric scenes. Follow-up works extended this idea to real-world scenes, achieving impressive zero-shot generalization. Notably, ZeroNVS~\cite{zeronvs} fine-tunes a diffusion model on large-scale real-world scenes, focusing on single-frame reconstruction. Reconfusion~\cite{wu2023reconfusion} trains a conditional diffusion model in combination with a PixelNeRF~\cite{pixelnerf} objective for more accurate conditioning. CAT3D~\cite{gao2024cat3d} trains a multi-view conditional diffusion model along with a feed-forward 3D reconstruction pipeline to render 3D consistent novel views. 

\noindent\textbf{Multi-View Depth Estimation.} Earlier works~\cite{packnet-semguided, monodepth2, tri-depthformer, tri-self_calibration} concentrated on in-domain depth estimation, which overfits to the specific camera geometry and requires similar training and testing domains. Most recently, zero-shot monocular depth estimation~\cite{midas, omnidata, zoedepth, tri-zerodepth} has gained significant interest, where models are trained on large-scale mixture of datasets and demonstrate impressive generalization to out-of-domain inference without finetuning. Notably,~\cite{midas, omnidata, metric3d, hu2024metric3d, birkl2023midasv31model, depthanything, depth_anything_v2} scale up their training to large-scale real or a mix of real and synthetic datasets. Recent works have also used priors from 2D diffusion models as denoisers~\cite{marigold, gui2024depthfm,fu2024geowizard} for monocular depth estimation, as well as explored video~\cite{luo2020consistent,shao2024learning,DepthCrafter} and multi-view settings~\cite{deep_mv_depth,yao2018mvsnet,hu2024mvdfusion}. Notably, very few works have focused on the task of novel depth synthesis~\cite{tri-define}, relying on intermediate 3D representations to generate scene geometry.

\section{Multi-View Geometric Diffusion}

Given a collection $\mathcal{I}_C = \{\textbf{I},\mathcal{C}\}_{n=1}^N$ of input images $\textbf{I}_n \in \mathbb{R}^{H \times W \times 3}$ and corresponding cameras $\mathcal{C}_n = \{\textbf{K},\textbf{T}\}$ with intrinsics $\textbf{K} \in \mathbb{R}^{3 \times 3}$ and extrinsics $\textbf{T} \in \mathbb{R}^{4 \times 4}$, our goal is to generate a predicted image $\hat{\textbf{I}}_t \in \mathbb{R}^{H \times W \times 3}$ and depth map $\hat{\textbf{D}}_t \in \mathbb{R}^{H \times W}$ for a novel target camera $\mathcal{C}_t$. We use a diffusion model $f_\theta \sim p(\hat{\textbf{I}}_t,\hat{\textbf{D}}_t|\mathcal{C}_t,\mathcal{I}_C)$ to learn a conditional distribution from which to sample novel target images and depth maps. A diagram of our proposed Multi-View Geometric Diffusion (\Acronym) architecture is shown in Figure~\ref{fig:diagram}, and below we describe each of its components in detail.

\subsection{Preliminaries}

\subsubsection{Diffusion Overview}
Diffusion models~\cite{sohl2015deep, ho2020denoising, chang2023design} operate by learning a state transition function from a noise tensor $\bm{\epsilon}$ to a sample $\textbf{x}_0$ from a learned data distribution. 
Such function is defined as:
\begin{equation}
    \textbf{x}_t = \sqrt{\alpha_t} \textbf{x}_0 + \sqrt{1 - \alpha_t}\bm{\epsilon}
\label{eq:velocity}
\end{equation}
where $\bm{\epsilon} \sim \mathcal{N}(\mathbf{0}, \mathbb{I})$, $\alpha_t = \prod_{s=1}^t \left( 1 - \beta_s \right)$ and $\{\beta_t\}_{t=1}^T$ is the variance schedule for a process with $T$ steps.
A neural network $\hat{\bm{\epsilon}} = f_\theta(\textbf{x}_t,t,\textbf{c})$ is trained to estimate the noise $\hat{\bm{\epsilon}}_t$ added to a sample $\textbf{x}_0$ at timestep $t$, given a conditioning variable $\textbf{c}$ used to control the generative process~\cite{NEURIPS2021_49ad23d1,rombach2022high,liu2023zero1to3}. 
At inference time, a novel $\textbf{x}_0$ is reconstructed from a normally-distributed variable $\textbf{x}_T \sim \mathcal{N}\left(\textbf{0},\mathbb{I}\right)$ by iteratively applying the learned transition function $f_\theta$ over $T$ steps.

A key choice when designing diffusion models is the architecture used to parameterize $f_\theta$. 
Initial implementations relied on convolutional architectures~\cite{ronneberger2015u,rombach2022high,ho2022cascaded,NEURIPS2021_49ad23d1}, due to their simplicity and easiness to train. 
At high resolutions this becomes costly,
which led to the popularization of latent diffusion, in which $f_\theta$ is not trained in pixel space, but in a lower-dimensional latent space of an image auto-encoder~\cite{NIPS2017_7a98af17}. 
Although more efficient, this approach has its drawbacks: the loss of fine-grained details due to compression, and the need for dense 2D grid-like inputs. %

\subsubsection{RIN Architecture}

Transformer-based architectures~\cite{Peebles2022DiT} have been explored as a more scalable alternative that does not require structured grid-like inputs. 
However, the squared complexity of attention still poses a challenge at higher resolutions. 
Hence, in this work we use Recurrent Interface Networks (RIN)~\cite{rin} as a more efficient Transformer-based architecture.
The key idea is the separation of computation into input tokens $\textbf{X} \in \mathbb{R}^{N \times D}$ and latent tokens $\textbf{Z} \in \mathbb{R}^{L \times D}$, where the former is obtained by tokenizing input data (and thus depends on the input size $N$), but $L$ is a fixed dimension.
At each RIN block, the latents $\textbf{Z}$ are first cross-attended with the inputs $\textbf{X}$, followed by several self-attention layers on $\textbf{Z}$, and the resulting latents are cross-attended back with $\textbf{X}$. 

The fact that the bulk of our computation (i.e., self-attention) operates on a fixed number of $L$ latent tokens, rather than on all $N$ input tokens, makes it affordable to learn $f_\theta$ directly in pixel space. %
It also enables the use of significantly more conditioning views to generate scene tokens (as shown in the supplementary material).
Moreover, we show that RIN latent tokens can be incrementally expanded to allow the training of larger models by fine-tuning smaller ones, with promising scaling behavior in terms of performance versus complexity (Sec. \ref{sec:scaling}).

\subsection{Scene Scale Normalization}
\label{sec:rpn}

Scale estimation is an inherently ambiguous problem in 3D
computer vision. Single-frame methods are unable to directly estimate scale, and must rely on other sources to disambiguate~\cite{packnet,tri-zerodepth,fsm,metric3d,unidepth,grin}. 
Multi-frame methods must inherit the scale given by cross-camera extrinsics, to preserve multi-view geometry. 
This is particularly challenging when training across multiple datasets, that use different calibration procedures, such as metric scale from additional sensors (e.g., LiDAR, IMUs, or stereo pairs) or arbitrary scale from self-supervision (e.g., COLMAP~\cite{schonberger2016structure}). 
Several works have explored ways to address this challenge, such as scale jittering to increase robustness~\cite{tri-define},  
scale conditioning~\cite{chan2023genvs}, or test-time multi-view alignment~\cite{charatan23pixelsplat}.

We propose a simple 
technique designed to automatically extract scene scale from conditioning cameras and ground-truth depth, by normalizing the input to our diffusion process. This scene scale is injected back to generated predictions, resulting in \emph{multi-view consistent} depth maps. %
First, all conditioning extrinsics $\textbf{T}_c^n$ are expressed relative to the novel target extrinsics $\textbf{T}_t$, so that $\tilde{\textbf{T}}_c^n = \textbf{T}_c^n \textbf{T}_t^{-1}$. Note that this means that the novel normalized target camera $\tilde{\mathcal{C}}_t = \{\textbf{K},\tilde{\textbf{T}}\}_t$ is always positioned at the origin. This enforces translational and rotational invariance to scene-level coordinate changes, a property that has been shown to improve multi-view depth estimation~\cite{tri-define,epio}. %

We define the scene scale $s$ as a scalar representing the largest absolute camera translation in any spatial coordinate, i.e., $s = \max \{ \{ |\tilde{x}|,|\tilde{y}|,|\tilde{z}| \}_c^n  \}_{n=1}^{N}$, where $\textbf{t}_c^n = \left[x,y,z\right]^T$ is the translation component of
$\textbf{T}_c^n = \scriptsize \begin{bmatrix}
    \textbf{R} & \textbf{t} \\
    \textbf{0} & 1      
\end{bmatrix}$, and $\textbf{R}_c^n \in \mathbb{R}^{3 \times 3}$ is its rotational component. We then divide all translation vectors by this value, such that $\tilde{\textbf{t}}_c^n = \left[\nicefrac{x}{s},\nicefrac{y}{s},\nicefrac{z}{s}\right]^T$. 
During training, if the target depth map $\textbf{D}_t$ is used as ground-truth, we also divide it by $s$ to maintain the scene geometry consistent across views, such that $\tilde{\textbf{D}}_t = \nicefrac{\textbf{D}_t}{s}$. If $\max \{\tilde{\textbf{D}}_t\} > d_{max}$ (the maximum value estimated by our model), we recalculate the scene scale as $s' = s \cdot \nicefrac{D_{max}}{\max \{ \tilde{\textbf{D}}_t \} }$ so the normalized ground-truth is within range, and use this new value to recalculate $\{\textbf{t}_c^n\}_{n=1}^N$. %
During inference, once $\hat{\textbf{D}}_t$ is generated, we multiply it by $s$ to ensure consistency with conditioning cameras. In other words, \emph{generated depth maps will have the same scale as conditioning cameras}. 

\subsection{Conditioning Embeddings}
\label{sec:emb}

\noindent\textbf{Image Encoder.} 
We use EfficientViT~\cite{cai2022efficientvit} to tokenize input conditioning views, providing visual scene information for novel generation. We start from a pre-trained \texttt{EfficientViT-SAM-L2} model taken from the official repository, and fine-tune it end-to-end during training. A $H \times W$ input image $\textbf{I}$ will result in $\textbf{F}_{\textbf{I}} \in \mathbb{R}^{\frac{H}{4} \times \frac{W}{4} \times 448}$ features. These features are flattened and processed by a linear layer $\mathcal{L}^I_{448 \rightarrow D_I}$ to produce image embeddings $E_c^{I,n} \in \mathbb{R}^{\frac{HW}{16} \times D_I}$. This process is repeated for each conditioning view, resulting in $N$ sets of image embeddings.

\noindent\textbf{Ray Encoder.} 
We use Fourier encoding~\cite{mildenhall2020nerf,tri-define,tri-zerodepth} to tokenize input cameras, parameterized as a raymap containing origin $\textbf{t}_{ijk} = \left[x,y,z\right]_k^T$ and viewing direction
$\textbf{r}_{ijk} = \left( \textbf{K}_k 
\textbf{R}_k\right)^{-1}\left[u_{ij},v_{ij},1 \right]^T$ for each pixel $\textbf{p}_{ij}$ from camera $k$. 
This information is used to (a) position features extracted from conditioning views in 3D space; %
and (b) determine novel viewpoints for image and depth synthesis. %
Conditioning cameras $\mathcal{C}_n$ are resized to match the resolution of image embeddings, and the target camera $\mathcal{C}_t$ is kept the same (note that $\textbf{t}_t$ is always at the origin, and $\textbf{R}_t = \mathbb{I}$). 
Assuming $N_o$ and $N_r$ origin and ray frequencies, the resulting ray embeddings are of dimensionality $D_{R} =
3 \left( N_o + N_r + 1\right).$

\subsection{Multi-Task Learning}
\label{sec:task}

Differently from other methods~\cite{pixelnerf,wu2023reconfusion,gao2024cat3d,charatan23pixelsplat,chen2024mvsplat}, \Acronym does not maintain an intermediate 3D representation. These methods rely on such representations to produce multi-view consistent predictions, by either conditioning a NeRF on available images for novel rendering, or projecting Gaussians generated from available images onto novel viewpoints. Instead, we propose to generate novel renderings directly from our implicit model, which therefore must itself be multi-view consistent. We achieve that by jointly learning \emph{novel view and depth synthesis}, i.e., by directly rendering depth maps from novel viewpoints alongside images. %

A simple approach would be to generate RGB-D predictions as a single task. However, that would limit our training datasets to only those with dense ground-truth depth. An alternative would be to condition the latent tokens themselves, by appending task-specific tokens. However, that would (a) make it impossible to jointly generate predictions for both tasks, and (b) promote the separation of appearance and geometric priors, by creating dedicated latent tokens portions of the latent space. 
Thus, we propose to use learnable \emph{task embeddings} $\textbf{E}^{task} \in \mathbb{R}^{D_{task}}$ to guide each individual generation towards a specific task. The model's predictions are parameterized as follows, depending on the task: 

\noindent\textbf{RGB Parameterization.} Our pixel-level diffusion does not require latent auto-encoders, and therefore ground-truth images are simply normalized to $[-1,1]$ with $\textbf{P}_{RGB} = \left( \textbf{I} + 1 \right) / 2$. Generated predictions are converted back to images using the inverse operation $\hat{\textbf{I}} = 2\, \hat{\textbf{P}}_{RGB} + 1$.

\noindent\textbf{Depth Parameterization.} To preserve multi-view consistency, our generated depth predictions must be \emph{scale-aware}, and thus should cover a wide range of possible values. 
We assume $d_{min} = 0.1$ and $d_{max} = 200$, which makes \Acronym suitable for indoor and outdoor scenarios. Note that these values are not metric, since they are considered after scene scale normalization. %
We use log-scale parameterization~\cite{dmd,grin} (Equation \ref{eq:depth_enc}), and predictions are converted back using the inverse operation (Equation \ref{eq:depth_dec}).
\begin{align}
\vspace{-3mm}
\small
\textbf{P}_D =& 
2 \left( \log\left(
\frac{\textbf{D}}{s \cdot d_{min}} \right) / 
\log\left(
\frac{d_{max}}{d_{min}} 
\right)
\right) - 1
\label{eq:depth_enc}
\\
\hat{\textbf{D}} =& 
\exp \left(
\big(
2\, \hat{\textbf{P}}_D + 1
\big) \log\left( 
\frac{d_{max}}{d_{min}}
\right)
\right) d_{min} \cdot s
\label{eq:depth_dec} 
\end{align}

\subsection{Novel View and Depth Synthesis}
\label{sec:inference}

The steps described above produce two different sets of inputs: \emph{scene tokens}, used to contextualize the diffusion process; and \emph{prediction tokens}, used to guide the diffusion process towards generating the desired predictions. 

\noindent\textbf{Scene tokens} are obtained by first concatenating image and ray embeddings from each conditioning view, producing $\textbf{E}_{c}^{n} = \textbf{E}_c^{I,n} \oplus \textbf{E}_c^{R,n}$, and then concatenating embeddings from all conditioning views, producing $\textbf{E}_{c} = \textbf{E}_c^1 \oplus ... \oplus \textbf{E}_c^N \in \mathbb{R}^{\frac{NHW}{16} \times (D_I + D_R)}$. 
To improve training efficiency, we randomly sample $M_s$ scene tokens as conditioning. %

\noindent\textbf{Prediction tokens} are obtained by concatenating ray embeddings $\textbf{E}_t^R $ from the target camera with the desired task embeddings $\textbf{E}^{task}$ and \emph{state embeddings} $\textbf{S}_t^{task}$. The state embeddings contain the evolving state of the diffusion model's predictions, and are defined as follows:

\begin{itemize}

\item\textbf{Training:} State embeddings $\textbf{S}_t$ are generated by parameterizing 
an input image $\textbf{I}_t$ or depth map $\textbf{D}_t$, and adding random noise determined by a noise scheduler $n(t)$, given a randomly sampled timestep $t \in [1,T]$. 
Our diffusion model is trained to learn the transition function $f_\theta$ according to Equation \ref{eq:velocity}. We use L2 and L1 losses to supervise image and depth generation, respectively. For depth estimation, prediction tokens are generated only for pixels with valid ground-truth. 
For both tasks, to improve efficiency, we randomly sample $M_p$ prediction tokens. %

\item\textbf{Inference:} State embeddings $\textbf{S}_t^T \sim \mathcal{N}(\textbf{0},\mathbb{I})$ are sampled as 3-dimensional vectors for image or scalars for depth generation. 
They are iteratively denoised for $T$ steps using $f_\theta$ with scheduler $n(t)$. 
At $t=0$ state embeddings $\textbf{S}_t^0$ will contain the parameterized prediction, that is converted back to $\hat{\textbf{I}}_t$ or $\hat{\textbf{D}}_t$. 
To mitigate stochasticity, we perform test-time ensembling~\cite{marigold,grin} over $E=5$ samples.

\end{itemize}

\subsection{Incremental Multi-View Generation}
\label{sec:autoreg}

Due to stochasticity in the diffusion process, generating multiple predictions from unobserved areas may lead to inconsistencies, even though each one is equally valid. 
To address this ambiguity, we propose to maintain a history of generated images $\mathcal{I}_G = \{\hat{\textbf{I}},\mathcal{C}\}_{g=1}^G$ to use as additional conditioning for future generations. 
Although this approach creates additional scene tokes, the number of RIN latent tokens %
remain the same, which greatly increases efficiency. 
In fact, we show in the supplementary material that \Acronym can scale to thousands of views on a single GPU, leading to further improvements without any additional information. %

\vspace{-1mm}
\subsection{Fine-Tuning Larger Models}

The fixed dimensionality of our latent tokens $\textbf{Z}$ enables efficient training and inference in terms of the number of input tokens $\textbf{X}$. 
However, this bottleneck representation may also hinder the quality of generated samples, especially when considering large-scale multi-task learning. 
If this hypothesis is correct, increasing the number of latents tokens should lead to improvements, albeit at the cost of slower training times. 
To avoid training larger models from scratch, we observe that introducing more latent tokens does not change our architecture at all, i.e., the cross-attention with inputs and the self-attention between latents remains the same. Thus, after training with a specific number of latent tokens, we can simply duplicate and concatenate them, resulting in a structurally similar representation with twice the capacity.
This larger model can then be further optimized, and we empirically demonstrate (Sec. \ref{sec:scaling}) that this incremental fine-tuning strategy leads to substantial improvements without the need to train new models from scratch.

\begin{figure*}[t!]
\vspace{-5mm}
\begin{center}
    \centering
\includegraphics[width=0.24\textwidth,height=3.1cm]{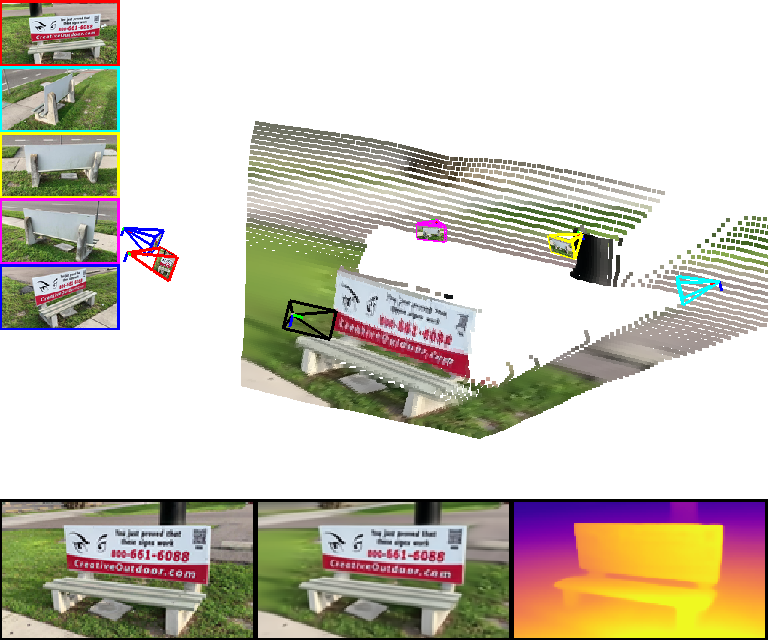}
\includegraphics[width=0.24\textwidth,height=3.1cm]{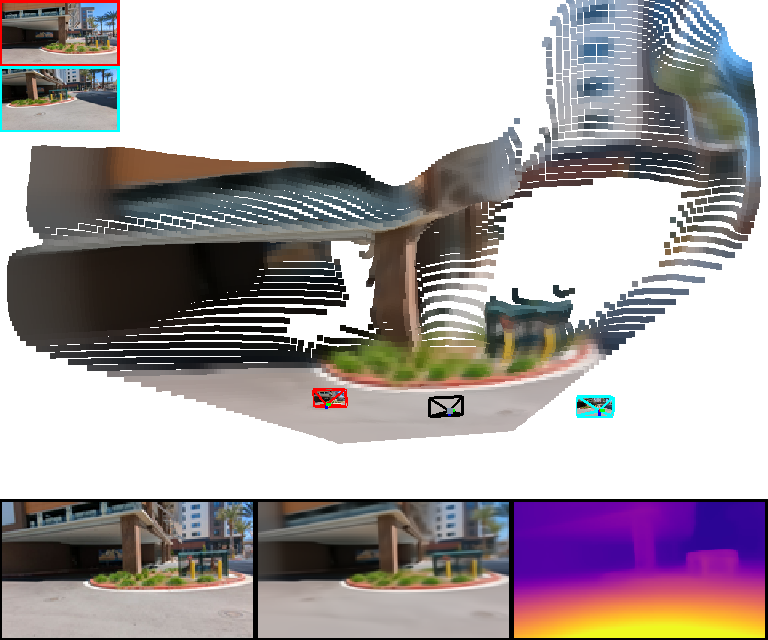}
\includegraphics[width=0.24\textwidth,height=3.1cm]{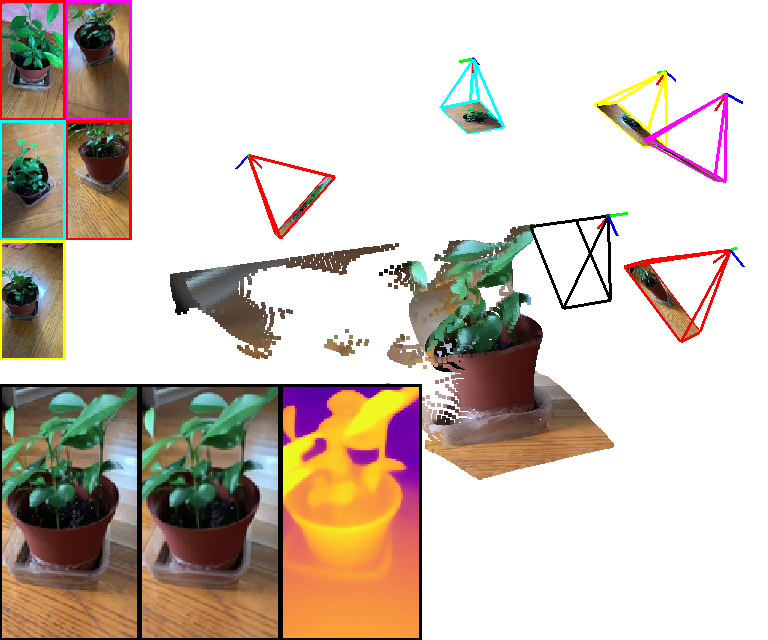}
\includegraphics[width=0.24\textwidth,height=3.1cm]{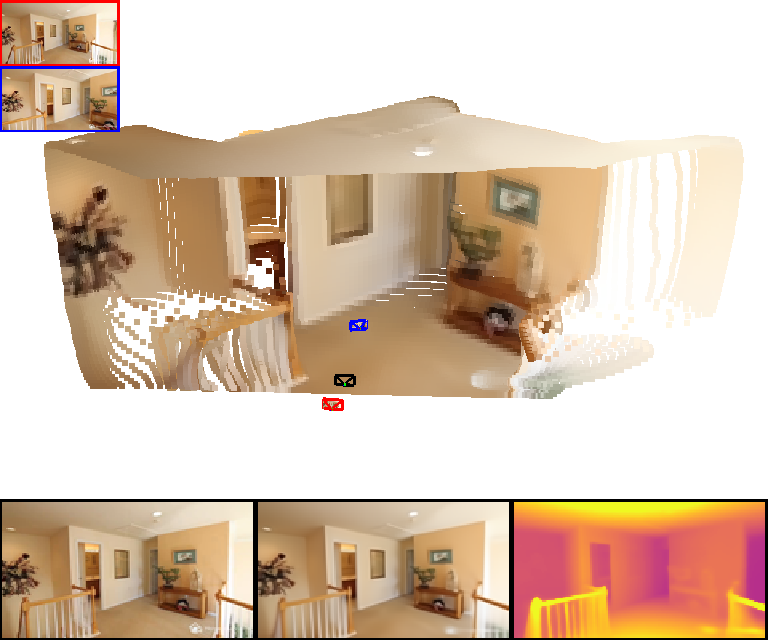}
\includegraphics[width=0.24\textwidth,height=3.1cm]{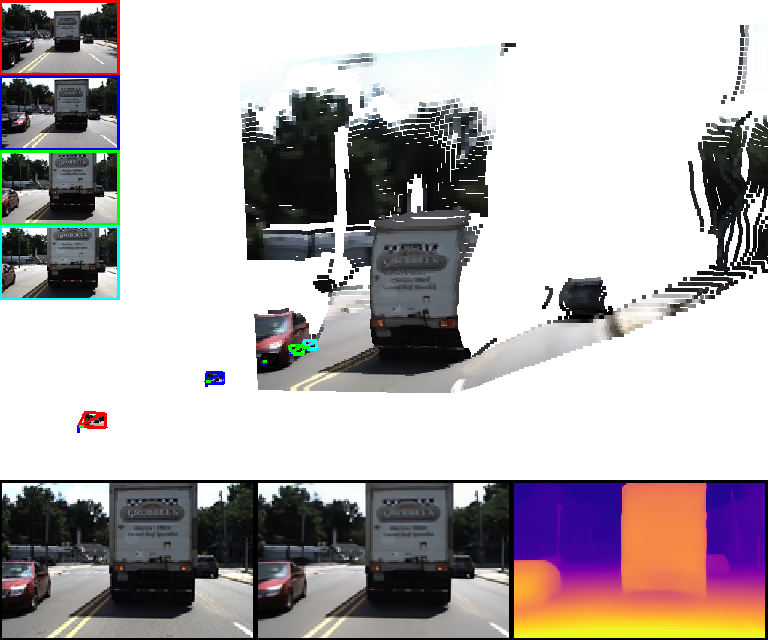}
\includegraphics[width=0.24\textwidth,height=3.1cm]{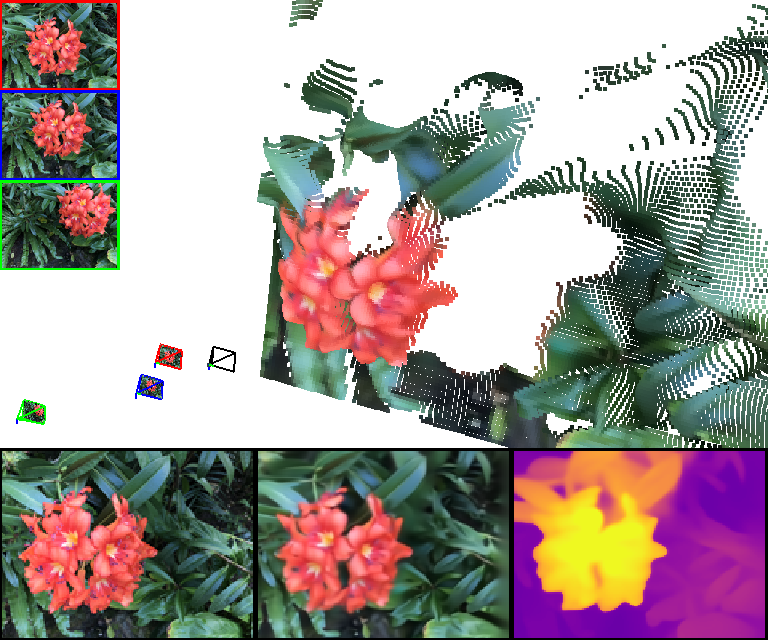}
\includegraphics[width=0.24\textwidth,height=3.1cm]{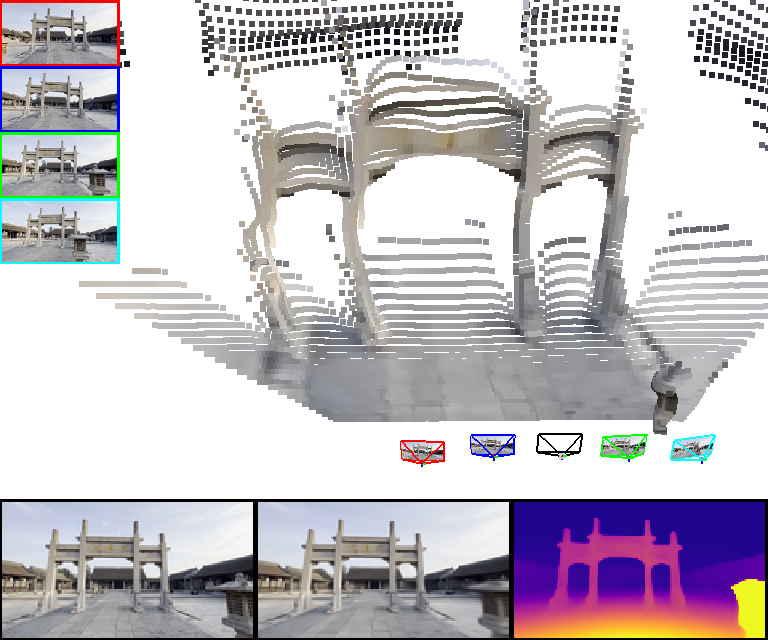}
\includegraphics[width=0.24\textwidth,height=3.1cm]{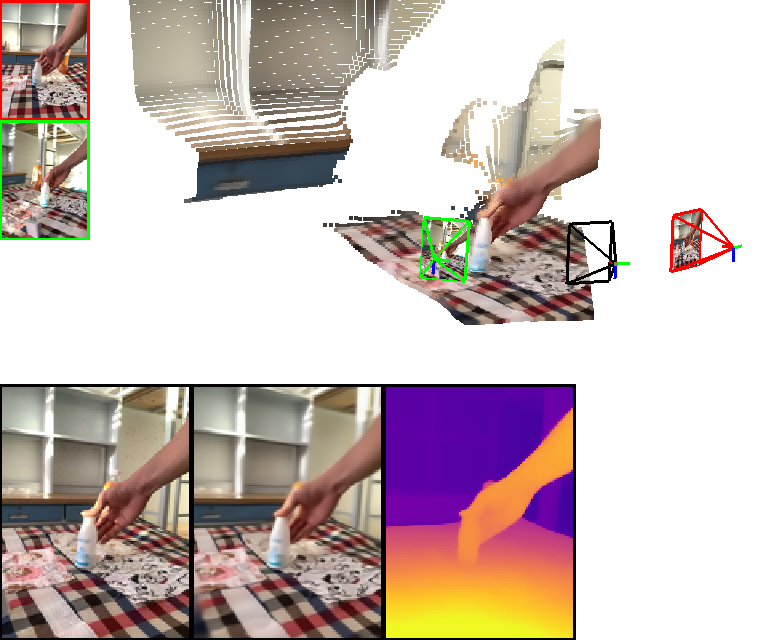}
\vspace{-2mm}
\caption{
    \textbf{\Acronym novel view and depth synthesis results} randomly sampled from different evaluation benchmarks and in-the-wild datasets. Top images are conditioning views (colored cameras), and bottom images are the target view (black camera), showing from left-to-right: ground-truth image, predicted image, and predicted depth map. These predictions are used to produce a colored 3D pointcloud observed from the target view. For more examples and additional visualizations, please refer to the supplementary material.
}
\label{fig:pointclouds}
\end{center}
\vspace{-9mm}
\end{figure*}

\vspace{-1mm}
\section{Experiments}

\vspace{-1mm}
\subsection{Architecture Details}

Our diffusion model is based on the RIN implementation from ~\cite{wang_rin_torch_2022}, containing $6$ blocks with depth $4$, $16$ latent and read-write heads, and $\textbf{Z} \in \mathbb{R}^{256 \times 1024}$. 
Ray embeddings were calculated using $N_o = N_r = 8$ and  maximum frequency $100$.  
Task embeddings are of dimensionality $D_{task} = 128$, and $D_I=256$.
At training time, we randomly select $M_s=1024$ scene tokens from each conditioning view, and $M_t = 4096$ prediction tokens. 
Prediction tokens are dedicated either for only image or depth generation, or split equally between both tasks. We select between $2-5$ views for conditioning, and an additional view as target. 
We use DDIM~\cite{ddim} with $1000$ training and $10$ evaluation timesteps, \emph{sigmoid} noise schedule, and EMA~\cite{kingma2014adam} with $\beta=0.999$. 
Images are resized to a longest side of $256$. %

We trained \Acronym from scratch, using AdamW~\cite{loshchilov2019decoupled} with $\beta_1=0.9$, $\beta_2=0.99$, batch size $b=512$, weight decay $w=10^{-2}$, learning rate $lr=10^{-4}$, and a warm-up scheduler~\cite{goyal2018accurate} with linear increase for $10$k steps followed by cosine decay, for a total of $300$k steps (until convergence). For each fine-tuning stage, we duplicate $\textbf{Z}$ and train for $50$k steps, using the same scheduler and half the learning rate. In total, training takes around $6$k A$100$ GPU-hours. Code and pre-trained models will be made available upon publication.

\subsection{Training Datasets}

Rather than relying on a intermediate 3D representation to generate novel predictions, \Acronym samples directly from an implicit latent space. 
To promote the learning of the priors required to achieve this, 
we curated more than 60M samples from publicly available multi-view datasets, focusing on diversity both in terms of appearance (e.g., indoor, outdoors, driving, robotics, objects, synthetic) and geometry (e.g., intrinsics, resolution, scale). 
Table \ref{tab:datasets} shows a list of the datasets used to train \Acronym, and more information can be found in the supplementary material. 
For a fair comparison with similar methods on in-domain benchmarks, wherever possible we use only samples from the official training splits. 
We also note that some datasets contain dynamic objects, which are currently not modeled by our framework. 
We decided to include these to further increase the diversity and zero-shot capabilities of \Acronym, and observed non-trivial improvements across all benchmarks (Table \ref{tab:ablation}). %

\subsection{Novel View Synthesis}

We evaluated \Acronym for the task of novel view synthesis with an arbitrary number of conditioning views, from $2$ to $9$. 
The $2$-view setting is dominated by 3DGS ~\cite{charatan23pixelsplat,chen2024mvsplat}, in which predictions from available views are scale-aligned and rendered from novel viewpoints; whereas settings with more views are dominated by methods that first generate additional conditioning images with diffusion, and then use a generalizable NeRF for novel view rendering~\cite{wu2023reconfusion,gao2024cat3d}. 

\begin{table}
\scriptsize
\renewcommand{\arraystretch}{0.9}
\setlength{\tabcolsep}{0.15em}
\centering
\begin{tabular}{l|ccc|ccc}
\toprule
\multirow{2}{*}{}
&
\multicolumn{3}{c|}{\emph{RealEstate10K}~\cite{zhou2018stereo}} &
\multicolumn{3}{c}{\emph{ACID}~\cite{infinite_nature_2020}} \\
&
PSNR$\uparrow$ &
SSIM$\uparrow$ &
LPIPS$\downarrow$ &
PSNR$\uparrow$ &
SSIM$\uparrow$ &
LPIPS$\downarrow$
\\
\midrule
PixelNeRF~\cite{pixelnerf}
& $20.43$ & $0.589$ & $0.550$  
& $20.97$ & $0.547$ & $0.533$
\\
GPNR~\cite{suhail2022generalizable}
& $24.11$ & $0.793$ & $0.255$  
& $25.28$ & $0.764$ & $0.332$  
\\
AttnRend~\cite{Du_2023_CVPR}
& $24.78$ & $0.820$ & $0.213$  
& $26.88$ & $0.799$ & $0.218$  
\\
MuRF~\cite{xu2024murf}
& \bestC{$26.10$} & \bestC{$0.858$} & $0.143$  
& $28.09$ & \bestC{$0.841$} & $0.155$  
\\
PixelSplat~\cite{charatan23pixelsplat}
& $25.89$ & \bestC{$0.858$} & \bestC{$0.142$}  
& \bestC{$28.14$} & $0.839$ & \bestC{$0.150$}  
\\
MVSplat~\cite{chen2024mvsplat}
& \bestB{$26.39$} & \bestB{$0.869$} & \bestB{$0.128$}  
& \bestB{$28.25$} & \bestB{$0.843$} & \bestB{$0.144$}  
\\
\midrule
\textbf{\Acronym}
& \bestA{$28.41$} & \bestA{$0.891$} & \bestA{$0.107$}  
& \bestA{$29.98$} & \bestA{$0.875$} & \bestA{$0.131$} 
\\
\bottomrule
\end{tabular}
\vspace{-2mm}
\caption{
\textbf{Novel view synthesis results} with 2 input views. 
}
\label{tab:nvs_2view}
\vspace{-5mm}
\end{table}

\begin{table*}
\scriptsize
\renewcommand{\arraystretch}{0.8}
\centering
\vspace{-5mm}
\begin{tabular}{l|l|ccc|ccc|ccc}
\toprule
& &
\multicolumn{3}{c|}{\emph{PSNR}$\uparrow$} &
\multicolumn{3}{c|}{\emph{SSIM}$\uparrow$}  &
\multicolumn{3}{c}{\emph{LPIPS}$\downarrow$} \\
\cmidrule(lr){3-5} 
\cmidrule(lr){6-8} 
\cmidrule(lr){9-11} 
& &
3-view & 6-view & 9-view &
3-view & 6-view & 9-view &
3-view & 6-view & 9-view
\\
\midrule
\multirow{8}{*}{\rotatebox{90}{\scriptsize{RE10K~\cite{zhou2018stereo}}}}
& FreeNeRF~\cite{Yang2023FreeNeRF}
& $20.54$ & $25.63$ & $27.32$   
& $0.731$ & $0.817$ & $0.843$    
& $0.394$ & $0.344$ & $0.332$   
\\
& SimpleNeRF~\cite{somraj2023simplenerf}
& $23.89$ & $28.75$ & $29.55$   
& $0.839$ & $0.896$ & $0.900$   
& $0.292$ & $0.239$ & $0.236$    
\\
& ZeroNVS$^*$\cite{zeronvs}
& $19.11$ & $22.54$ & $23.73$   
& $0.675$ & $0.744$ & $0.766$   
& $0.422$ & $0.374$ & $0.358$   
\\
& ReconFusion$^\dagger$~\cite{wu2023reconfusion}
& \bestC{$25.84$} & \bestC{$29.99$} & \bestC{$31.82$}   
& \bestC{$0.910$} & \bestC{$0.951$} & \bestC{$0.961$}
& \bestC{$0.144$} & \bestC{$0.103$} & \bestC{$0.092$}   
\\
& CAT3D$^\dagger$~\cite{gao2024cat3d}
& \bestB{$26.78$} & \bestB{$31.07$} & \bestB{$32.20$}   
& \bestB{$0.917$} & \bestB{$0.954$} & \bestB{$0.963$}   
& \bestB{$0.132$} & \bestB{$0.092$} & \bestA{$0.082$}
\\
\cmidrule(lr){2-11} 
& \textbf{\Acronym}
& \bestA{$28.70$} & \bestA{$31.71$} & \bestA{$32.89$}   
& \bestA{$0.925$} & \bestA{$0.960$} & \bestA{$0.969$}   
& \bestA{$0.128$} & \bestA{$0.090$} & \bestB{$0.085$}   
\\
\midrule
\multirow{8}{*}{\rotatebox{90}{\scriptsize{LLFF~\cite{mildenhall2019llff}}}}
& FreeNeRF~\cite{Yang2023FreeNeRF}
& $19.63$ & $23.72$ & $25.12$   
& $0.613$ & $0.773$ & $0.820$   
& $0.347$ & $0.232$ & $0.193$  
\\
& SimpleNeRF~\cite{somraj2023simplenerf}
& $19.24$ & $23.05$ & $23.98$   
& $0.623$ & $0.737$ & $0.762$   
& $0.375$ & $0.296$ & $0.286$   
\\
& ZeroNVS$^*$\cite{zeronvs}
& $15.91$ & $18.39$ & $18.79$   
& $0.359$ & $0.449$ & $0.470$   
& $0.512$ & $0.438$ & $0.416$   
\\
& ReconFusion$^\dagger$~\cite{wu2023reconfusion}
& \bestC{$21.34$} & \bestC{$24.25$} & \bestC{$25.21$}   
& \bestC{$0.724$} & \bestC{$0.815$} & \bestC{$0.848$}   
& \bestC{$0.203$} & \bestC{$0.152$} & \bestC{$0.134$}
\\
& CAT3D$^\dagger$~\cite{gao2024cat3d}
& \bestB{$21.58$} & \bestB{$24.71$} & \bestB{$25.63$}   
& \bestB{$0.731$} & \bestB{$0.833$} & \bestA{$0.860$} 
& \bestB{$0.181$} & \bestA{$0.121$} & \bestB{$0.107$}   
\\
\cmidrule(lr){2-11} 
& \textbf{\Acronym}
& \bestA{$22.39$} & \bestA{$24.76$} & \bestA{$25.93$}  
& \bestA{$0.776$} & \bestA{$0.882$} & \bestB{$0.853$}
& \bestA{$0.157$} & \bestB{$0.131$} & \bestA{$0.103$}
\\
\midrule
\multirow{8}{*}{\rotatebox{90}{\scriptsize{DTU~\cite{jensen2014large}}}}
& FreeNeRF~\cite{Yang2023FreeNeRF}
& $20.46$ & $23.48$ & \bestC{$25.56$}   
& $0.826$ & $0.870$ &        $0.902$   
& $0.173$ & $0.131$ &        $0.102$    
\\
& SimpleNeRF~\cite{somraj2023simplenerf}
& $16.25$ & $20.60$ & $22.75$   
& $0.751$ & $0.828$ & $0.856$   
& $0.249$ & $0.190$ & $0.176$   
\\
& ZeroNVS$^*$\cite{zeronvs}
& $16.71$ & $17.70$ & $17.92$   
& $0.716$ & $0.737$ & $0.745$   
& $0.223$ & $0.205$ & $0.200$   
\\
& ReconFusion$^\dagger$~\cite{wu2023reconfusion}
& \bestC{$20.74$} & \bestC{$23.62$} &        $24.62$   
& \bestB{$0.875$} & \bestB{$0.904$} & \bestC{$0.921$}   
& \bestC{$0.124$} & \bestC{$0.105$} & \bestC{$0.094$}   
\\
& CAT3D$^\dagger$~\cite{gao2024cat3d}
& \bestB{$22.02$} & \bestB{$24.28$} & \bestB{$25.92$}   
& \bestC{$0.844$} & \bestC{$0.899$} & \bestB{$0.928$}   
& \bestB{$0.121$} & \bestB{$0.095$} & \bestA{$0.073$}   
\\
\cmidrule(lr){2-11} 
& \textbf{\Acronym}
& \bestA{$23.79$} & \bestA{$25.31$} & \bestA{$26.88$}   
& \bestA{$0.893$} & \bestA{$0.912$} & \bestA{$0.939$}  
& \bestA{$0.115$} & \bestA{$0.091$} & \bestB{$0.088$}    
\\
\midrule
\multirow{8}{*}{\rotatebox{90}{\scriptsize{CO3D~\cite{reizenstein21co3d}}}}
& FreeNeRF~\cite{Yang2023FreeNeRF}
& $13.28$ & $15.20$ & $17.35$   
& $0.461$ & $0.523$ & $0.575$   
& $0.634$ & $0.596$ & $0.561$   
\\
& SimpleNeRF~\cite{somraj2023simplenerf}
& $15.40$ & $18.12$ & $20.52$   
& $0.553$ & $0.622$ & $0.672$   
& $0.612$ & $0.541$ & $0.493$   
\\
& ZeroNVS$^*$\cite{zeronvs}
& $17.13$ & $19.72$ & $20.50$  
& $0.581$ & $0.627$ & $0.640$  
& $0.566$ & $0.515$ & $0.500$   
\\
& ReconFusion$^\dagger$~\cite{wu2023reconfusion}
& \bestC{$19.59$} & \bestC{$21.84$} & \bestC{$22.95$}   
& \bestC{$0.662$} & \bestC{$0.714$} & \bestC{$0.736$}   
& \bestC{$0.398$} & \bestC{$0.342$} & \bestC{$0.318$}    
\\
& CAT3D$^\dagger$~\cite{gao2024cat3d}
& \bestB{$20.57$} & \bestB{$22.79$} & \bestB{$23.58$}   
& \bestB{$0.666$} & \bestB{$0.726$} & \bestB{$0.752$}   
& \bestB{$0.351$} & \bestA{$0.292$} & \bestB{$0.273$}   
\\
\cmidrule(lr){2-11} 
& \textbf{\Acronym}
& \bestA{$20.68$} & \bestA{$23.49$} & \bestA{$24.77$}   
& \bestA{$0.678$} & \bestA{$0.743$} & \bestA{$0.789$}   
& \bestA{$0.331$} & \bestB{$0.322$} & \bestA{$0.248$}   
\\
\midrule
\multirow{8}{*}{\rotatebox{90}{\scriptsize{MIP-NeRF360~\cite{barron2022mipnerf360}}}}
& FreeNeRF~\cite{Yang2023FreeNeRF}
& $12.87$ & $13.35$ & $14.59$   
& $0.260$ & $0.283$ & $0.319$    
& $0.715$ & $0.717$ & $0.695$   
\\
& SimpleNeRF~\cite{somraj2023simplenerf}
& $13.27$ & $13.67$ & $15.15$   
& $0.283$ & $0.312$ & $0.354$ 
& $0.741$ & $0.721$ & $0.676$   
\\
& ZeroNVS$^*$\cite{zeronvs}
& $14.44$ & $15.51$ & $15.99$   
& $0.316$ & $0.337$ & $0.350$   
& $0.680$ & $0.663$ & $0.655$   
\\
& ReconFusion$^\dagger$~\cite{wu2023reconfusion}
& \bestC{$15.50$} & \bestC{$16.93$} & \bestC{$18.19$}  
& \bestC{$0.358$} & \bestC{$0.401$} & \bestC{$0.432$}    
& \bestC{$0.585$} & \bestC{$0.544$} & \bestC{$0.511$}   
\\
& CAT3D$^\dagger$~\cite{gao2024cat3d}
& \bestB{$16.62$} & \bestB{$17.72$} & \bestB{$18.67$}   
& \bestB{$0.377$} & \bestB{$0.425$} & \bestB{$0.460$}   
& \bestB{$0.515$} & \bestA{$0.451$} & \bestA{$0.439$}
\\
\cmidrule(lr){2-11} 
& \textbf{\Acronym}
& \bestA{$18.74$} & \bestA{$20.28$} & \bestA{$21.18$} 
& \bestA{$0.425$} & \bestA{$0.463$} & \bestA{$0.512$}
& \bestA{$0.499$} & \bestB{$0.512$} & \bestB{$0.488$}  
\\
\bottomrule
\end{tabular}
\vspace{-2mm}
\caption{
\textbf{Novel view synthesis results} with $3$-$9$ conditioning views. 
$^*$ indicate methods fine-tuned and reported by \cite{wu2023reconfusion}. 
$^\dagger$ indicate the use of image diffusion models pre-trained with internal datasets. 
Colored numbers represent \bestA{first}, \bestB{second}, and \bestC{third} best ranked metrics. 
RealEstate10k and CO3D results are \emph{in-domain}, and results on other benchmarks are \emph{zero-shot}.
}
\label{tab:nvs_mview}
\vspace{-5mm}
\end{table*}

\begin{table}
\scriptsize
\renewcommand{\arraystretch}{0.8}
\centering
\begin{tabular}{l|l|ccc}
\toprule
Dataset & Split
& Abs.Rel.$\downarrow$ & RMSE$\downarrow$ & $\delta_{1.25}$$\uparrow$ 
\\
\toprule
\multirow{3}{*}{LLFF} 
& 3-view & $0.105$ & $8.378$ & $0.865$ 
\\
& 6-view & $0.098$ & $8.260$ & $0.879$ 
\\
& 9-view & $0.093$ & $8.006$ & $0.891$ 
\\
\midrule
\multirow{3}{*}{MIPNeRF360} 
& 3-view & $0.143$ & $2.445$ & $0.842$ 
\\
& 6-view & $0.124$ & $2.292$ & $0.866$ 
\\
& 9-view & $0.109$ & $2.106$ & $0.883$ 
\\
\midrule
\multirow{3}{*}{CO3D} 
& 3-view & $0.101$ & $4.654$ & $0.874$ 
\\
& 6-view & $0.086$ & $4.436$ & $0.898$ 
\\
& 9-view & $0.079$ & $4.195$ & $0.914$ 
\\
\bottomrule
\end{tabular}
\vspace{-2mm}
\caption{
\textbf{Novel depth synthesis results} with $3$-$9$ conditioning views, relative to COLMAP~\cite{schonberger2016structure} ground-truth. No test-time alignment was used, which means that predicted depth maps share the same scale as the provided camera extrinsics. 
}
\label{tab:depth_mview}
\vspace{-6mm}
\end{table}

\begin{table}[t!]
\scriptsize
\setlength{\tabcolsep}{0.2em}
\renewcommand{\arraystretch}{0.8}
\centering
\begin{tabular}{c|l|c|c|cc|cc|c}
\toprule
& 
& \emph{RE10K(2)} 
& \emph{DTU(3)}
& \multicolumn{2}{c|}{\emph{CO3D(3)}}
& \multicolumn{2}{c|}{\emph{LLFF(3)}}
& \emph{ScanNet(2)}
\\
\toprule
& \textbf{Variation}
& PSNR %
& PSNR %
& PSNR %
& Abs.Rel %
& PSNR %
& Abs.Rel %
& Abs.Rel %
\\\midrule
A & \, 2048 (scratch)
& $22.09$ & $20.22$ & $16.25$ & $0.120$ & $19.11$ & $0.148$ & $0.088$
\\\midrule
B & $-$ RGB
& ----- & ----- & ----- & $0.132$ & ----- & $0.138$ & $0.089$
\\
C & $-$ Depth
& $21.46$ & $19.31$ & $16.39$ & ----- & $19.01$ & ----- & ----- 
\\ %
D & $-$ SSN
& $20.15$ & $17.98$ & $14.67$ & $0.175$ & $16.87$ & $0.224$ & $0.329$ 
\\
E & $-$ Dynamic
& $23.24$ & $19.51$ & $17.04$ & $0.141$ & $19.28$ & $0.143$ & $0.087$
\\\midrule
& \textbf{\Acronym (short)}
& $24.44$ & $20.16$ & $17.98$ & $0.127$ & $20.19$ & $0.131$ & $0.081$ 
\\\midrule
\midrule
F & $+$ RGB (B)
& $25.15$ & $21.11$ & $18.04$ & $0.120$ & $20.21$ & $0.120$ & $0.075$
\\
G & $+$ Depth (C)
& $25.68$ & $21.31$ & $18.39$ & $0.124$  & $20.52$ & $0.127$ & $0.079$ 
\\\midrule
& \textbf{\Acronym (full)}
& $25.89$ & $21.32$ & $18.48$ & $0.121$ & $20.58$ & $0.119$ & $0.075$ 
\\\bottomrule
\end{tabular}
\vspace{-3mm}
\caption{
\textbf{\Acronym ablation analysis.} Top results obtained with $200$k steps and $256$ latent tokens. Bottom results obtained by fine-tuning top models for $100$k steps. %
}
\vspace{-7mm}
\label{tab:ablation}
\end{table}

\begin{table*}[ht]
\vspace{-3mm}
\renewcommand{\arraystretch}{0.8}
\setlength{\tabcolsep}{0.15em}
\scriptsize
\centering
\parbox{.60\linewidth}{
    \centering
    \vspace{-2mm}
    \begin{tabular}{l|ccc|ccc|ccc}
\toprule
\multirow{2}{*}{\textbf{Method}} 
& \multicolumn{3}{c|}{\emph{ScanNet}~\cite{dai2017scannet}}
& \multicolumn{3}{c|}{\emph{SUN3D}~\cite{sun3d}}
& \multicolumn{3}{c}{\emph{RGB-D}~\cite{rgbdslam}}
\\ 
\cmidrule(lr){2-4} 
\cmidrule(lr){5-7} 
\cmidrule(lr){8-10} 
& 
Abs.Rel.$\downarrow$ & RMSE$\downarrow$ & $\delta_{1.25}\uparrow$ &
Abs.Rel.$\downarrow$ & RMSE$\downarrow$ & $\delta_{1.25}\uparrow$ &
Abs.Rel.$\downarrow$ & RMSE$\downarrow$ & $\delta_{1.25}\uparrow$ 
\\ \midrule
DPSNet~\cite{im2019dpsnet}
& $0.126$ & $0.315$ &   -   & $0.147$ & $0.449$ & $0.781$ & $0.151$ & $0.695$ & $0.804$ 
\\
NAS~\cite{kusupati2020normal}
& $0.107$ & $0.281$ &   -   & $0.127$ & $0.378$ & $0.829$ & $0.131$ & $0.619$ & $0.857$ 
\\
IIB~\cite{iib}
& $0.116$ & $0.281$ & $0.908$ & $0.099$ & $0.293$ & $0.902$ & $0.095$ & $0.550$ & $0.907$ 
\\
DeFiNe~\cite{tri-define}
& $0.093$ & $0.246$ & $0.911$ & $0.095$ & $0.287$ & $0.914$ & $0.095$ & $0.539$ & $0.909$ 
\\
EPIO~\cite{epio}
& \bestB{$0.086$} & \bestC{$0.229$} & \bestC{$0.923$} & \bestB{$0.090$} & \bestC{$0.260$} & \bestB{$0.912$} & \bestB{$0.080$} & \bestB{$0.433$} & \bestC{$0.912$} 
\\
GRIN~\cite{grin}
& \bestC{$0.088$} & \bestB{$0.224$} & \bestB{$0.925$} & \bestC{$0.097$} & \bestB{$0.274$} & \bestC{$0.908$} & \bestC{$0.092$} & \bestC{$0.512$} & \bestB{$0.919$} 
\\
\midrule
\textbf{\Acronym} 
& \bestA{$0.065$} & \bestA{$0.202$} & \bestA{$0.946$} & \bestA{$0.088$} & \bestA{$0.247$} & \bestA{$0.926$} & \bestA{$0.074$} & \bestA{$0.405$} & \bestA{$0.938$} 
\\
\bottomrule
\end{tabular}

    \vspace{-2mm}
    \caption{\textbf{Stereo depth experiments}, with 2 conditioning views. Results on ScanNet are \emph{in-domain}, and results on SUN3D and RGB-D are \emph{zero-shot}.
    }
    \label{tab:stereo}
}
\hfill 
\parbox{.35\linewidth}{
    \centering
    \vspace{-3mm}
    \begin{tabular}{l|ccc}
    \toprule
    \textbf{Method} &
    Abs.\ Rel$\downarrow$ &
    Sq.\ Rel$\downarrow$ &
    RMSE$\downarrow$ 
    \\
\midrule
DeMoN~\cite{ummenhofer2017demon} 
& $0.231$ & $0.520$ & $0.761$ 
\\
MiDas-v2~\cite{midas} 
& $0.208$ & $0.318$ & $0.742$ 
\\
BA-Net~\cite{tang2018ba} 
& $0.091$ & $0.058$ & $0.223$ 
\\
CVD~\cite{luo2020consistent} 
& $0.073$ & $0.037$ & $0.217$ 
\\
DeFiNe~\cite{tri-define}
& $0.059$ & \bestC{$0.022$} & $0.184$ 
\\
DeepV2D~\cite{deepv2d} 
& \bestC{$0.057$} & \bestB{$0.010$} & \bestC{$0.168$} 
\\
NeuralRecon~\cite{Sun_2021_CVPR} 
& \bestB{$0.047$} & $0.024$ & \bestB{$0.164$}
\\
\midrule
\textbf{\Acronym}
& \bestA{$0.041$} & \bestA{$0.007$} & \bestA{$0.143$} 
\\
\bottomrule
\end{tabular}

    \vspace{-2mm}
    \caption{\textbf{Video depth experiments} on ScanNet (\emph{in-domain}), with $10$ conditioning views.
    }
    \label{tab:video}
}
\end{table*}

\begin{table*}[t!]
\scriptsize
\vspace{-3mm}
\setlength{\tabcolsep}{0.35em}
\renewcommand{\arraystretch}{0.85}
\centering
    \begin{tabular}{r|r||cc||cc|cc||cc|cc||cc|cc}
        \toprule
        \multirow{2}{*}{\rotatebox{30}{\textbf{\# Latents}}}
        & \multirow{2}{*}{\rotatebox{30}{\textbf{\# Param.}}}
        & \multicolumn{2}{c||}{\emph{RE10K (2-view)}}
        & \multicolumn{4}{c||}{\emph{CO3Dv2 (3-view)}}
        & \multicolumn{4}{c||}{\emph{MIPNeRF360 (3-view)}}
        & \multicolumn{4}{c}{\emph{LLFF (3-view)}}
        \\
        \cmidrule(lr){3-4} 
        \cmidrule(lr){5-8} 
        \cmidrule(lr){9-12} 
        \cmidrule(lr){13-16} 
        & 
        &
        \hspace{1mm} PSNR$\uparrow$ &
        \hspace{1mm} SSIM$\uparrow$ &
        PSNR$\uparrow$ &
        SSIM$\uparrow$ &
        Abs.Rel.$\downarrow$ &
        RMSE$\downarrow$ &
        PSNR$\uparrow$ &
        SSIM$\uparrow$ &
        Abs.Rel.$\downarrow$ &
        RMSE$\downarrow$ &
        PSNR$\uparrow$ &
        SSIM$\uparrow$ &
        Abs.Rel.$\downarrow$ &
        RMSE$\downarrow$
        \\
\midrule

$256$ & $417.9$M %
& 25.89 & 0.841
& 18.48 & 0.567 & 0.121 & 5.347  
& 17.71 & 0.415 & 0.156 & 2.789 
& 20.58 & 0.706 & 0.119 & 8.807
\\
$512$ & $418.2$M %
& 26.33 & 0.859  
& 19.56 & 0.590 & 0.109 & 5.048 
& 18.15 & 0.429 & 0.149 & 2.581 
& 21.14 & 0.738 & 0.111 & 8.686
\\
$1024$ & $418.7$M %
& 27.73 & 0.870 
& 20.08 & 0.622 & 0.104 & 4.766  
& 18.49 & 0.444 & 0.147 & 2.499 
& 21.85 & 0.754 & 0.109 & 8.547 
\\
$2048$ & $419.7$M %
& 28.41 & 0.891   
& 20.68 & 0.678 & 0.101 & 4.654  
& 18.74 & 0.465 & 0.143 & 2.445
& 22.39 & 0.776 & 0.105 & 8.378
\\
\bottomrule
\end{tabular}
\vspace{-3mm}
\caption{
\textbf{Incremental fine-tuning experiments.} The first model ($256$) was trained from scratch for $300$k steps, and each subsequent model was obtained by duplicating the latents of the previous one and fine-tuning for $50$k steps.  Even though the total number of learnable parameters barely changes ($+0.5\%$), results improve by a significant margin (up to $+20\%$) across all benchmarks and tasks.
}
\vspace{-5mm}
\label{tab:scaling}
\end{table*}

Table \ref{tab:nvs_2view} reports $2$-view quantitative results, showing that \Acronym significantly outperforms the previous state of the art. 
We attribute this improvement to a combination of (a) our proposed large-scale training procedure, that leads to improvements even on in-domain benchmarks; and (b) the implicit learning of novel view and depth synthesis from such diverse data, instead of relying on indirect renderings.
We also note that these methods struggle with more than 2 views, since explicit geometric alignment between cameras is required. In contrast, \Acronym generates images and depth maps from novel viewpoints without any additional steps.

Therefore, we also evaluated the same \Acronym model on benchmarks with larger number of input views. %
Table \ref{tab:nvs_mview} reports these results, and once again we outperform strong baselines that also leverage large-scale pre-training in combination with diffusion models~\cite{wu2023reconfusion,gao2024cat3d}.
Note that \Acronym was trained using only $2$-$5$ conditioning views, and still it was able to generalize to larger numbers (additional experiments are available in the supplementary material).
For completeness, in Table \ref{tab:depth_mview} we report novel depth synthesis results compared to sparse COLMAP~\cite{schonberger2016structure} ground-truth, showing a similar trend of improvements with more conditioning views.
Qualitative results can be found in Figure \ref{fig:pointclouds}, including reconstructed pointclouds from novel viewpoints.

\subsection{Multi-View Depth Estimation}

To further evaluate the implicit geometric priors learned by \Acronym, we evaluated our model on two different multi-view depth estimation benchmarks. The \emph{Stereo} benchmark involves 2-view conditioning from overlapping image pairs, and the \emph{Video} benchmark involves $10$-view conditioning from image sequences. Note that in these benchmarks the target image is also used as conditioning, and the goal is to generate its depth map given all available views (a setting not considered during training). Results are reported in Tables \ref{tab:stereo} and \ref{tab:video}, where \Acronym once again outperforms the previous state of the art by a large margin, even though it does not rely on geometry-preserving 3D augmentations~\cite{tri-define}, architectural equivariance~\cite{epio}, or TSDF volumes~\cite{Sun_2021_CVPR}.

\subsection{Incremental Fine-Tuning}
\label{sec:scaling}

In Table \ref{tab:scaling} we analyze the impact that our proposed incremental fine-tuning strategy has on novel view and depth synthesis. 
The first noticeable aspect is that, even though adding more latent tokens barely changes the number of parameters, it leads to substantial improvements across all benchmarks. 
We attribute this to the use of latent tokens in the first place: even though it dramatically increases efficiency, computation at a lower-dimensional space hinders prediction quality. 
This is especially important in novel view synthesis, that benefits from high frequency details. 
However, experiments with direct self-attention~\cite{Peebles2022DiT} were not possible due to memory constraints: a $256 \times 256$ target image has $65536$ prediction tokens, plus $4096$ scene tokens per conditioning view. 
In contrast, our largest model has only $2048$ latent tokens, regardless of resolution or number of inputs. 
Moreover, we observed a sub-linear increase in inference time: a $256$ model generates predictions in roughly $0.3$s, while a $2048$ model ($8\times$) takes $1.1$s ($4\times$).

The second aspect is how efficiently smaller models can be improved with our proposed incremental fine-tuning strategy. 
Once the first model has been trained for $300$k steps, each subsequent one is fine-tuned for only $50$k steps. %
Similar to inference, we noticed that each training step of a $2048$ model is roughly $4\times$ slower compared to a $256$ model, which means that training our largest model from scratch would take $3\times$ longer compared to incremental fine-tuning. 
Moreover, we trained a $2048$ model from scratch for $300$k steps, and achieved worse performance than a $256$ model (\textbf{A} on Table \ref{tab:ablation}). This indicates that (a) training more complex models from scratch requires longer schedules, which further increases training time; and (b) our proposed strategy can leverage priors from smaller models and quickly adapt them to fit within a more complex representation.

\subsection{Ablation Study}
\label{sec:ablation}

Here we ablate our main design choices, with results summarized in Table \ref{tab:ablation}. 
For computational reasons, a shorter schedule of $200$k steps was used, achieving performance within $10\%$ of the full schedule. 

First, we analyzed \emph{single vs. multi-task learning, by training models capable of only novel (\textbf{B}) view or (\textbf{C}) depth synthesis.} 
As noted in~\cite{tri-define}, training a depth-only model led to slight degradation, in addition to limiting training datasets to only those with valid depth labels. 
However, an image-only model performed much worse, as evidence that geometric reasoning is key to accurate novel view synthesis. 

This hypothesis is reinforced when (\textbf{D}) \emph{training a multi-task model without scene scale normalization} (SSN), which led to significantly worse novel view and depth synthesis. 
Interestingly, novel view synthesis is worse without SSN than without multi-tasking, showing that inaccurate geometry is more detrimental than no geometry at all. 

We also validated the \emph{benefits of data diversity by training a model (\textbf{E}) without dynamic datasets}, which led to a decrease of roughly $12\%$ in the amount of training data. Even though \Acronym does not model dynamics, it still benefits from additional training data even on in-domain benchmarks. In the supplementary material we provide examples that show signs of implicit dynamics modeling. %

To further evaluate the multi-task capabilities of \Acronym, we experimented with the \emph{introduction of new tasks to a trained model, by fine-tuning (\textbf{F}) our depth-only model to include novel view synthesis, and (\textbf{G}) vice-versa}. 
The only added parameters were task embeddings for the new task, randomly initialized. 
This fine-tuning stage was performed for an $100$k steps, reaching our proposed full schedule for $256$ latents. Interestingly, the resulting 
model not only reached original task performance comparable to a multi-task model trained from scratch, but also on the added task, as indication that our learned multi-view priors can be repurposed for other tasks with minimal fine-tuning.%

\section{Conclusion}

We introduce \Acronym, a diffusion-based architecture for generalizable novel view and depth synthesis from an arbitrary number of posed images. We propose to generate novel images and depth maps by directly sampling from a multi-view consistent and scale-aware implicit distribution, jointly trained for both tasks. To achieve this, we curated over $60$M multi-view samples from publicly available datasets, and propose key technical modifications to the standard diffusion pipeline that enables efficient training in such diverse conditions. As a result, \Acronym achieves state-of-the-art results in multiple novel view synthesis benchmarks, as well as stereo and video depth estimation. We also design a novel incremental fine-tuning strategy to increase model complexity without retraining from scratch, showing promising scaling behavior with larger models. %

\appendix

\section{Datasets Preparation}

Each dataset was downloaded from its official source, following standard procedures.
Whenever possible, we used officially provided PyTorch dataloaders to access the relevant information (i.e., images, intrinsics, extrinsics, and depth maps), and wrote our own if not available. 
In all cases, data was mapped into a common format to enable mixed-batch training across all data sources. 
This includes padding images and intrinsics of different resolutions, and using empty depth maps whenever this label is not available. 
We treat video and multi-view datasets equally, by positioning all available cameras from the same scene on a global frame of reference, and considering all possible pairs as a potential source of training data. To select valid pairs for our purposes, we utilized three criteria, described below:  

\noindent\textbf{Camera center distance.} Conditioning views must have a camera center distance within $\textbf{t}_c^n$ within $c_{min} < ||\textbf{t}_c^n - \textbf{t}_t|| < c_{max}$, where $\textbf{t}$ is the camera's translation vector. In the case of dynamic datasets, we also apply the same constraint in a temporal sense to mitigate the impact of moving objects, such that $t_{min} < ||t_c^n - t_t|| < t_{max}$, where $t$ is the timestep of each camera within the sequence (fractional timesteps are used in the case of datasets with multiple asynchronized cameras). Assuming $c_M$ to be the maximum distance across any two cameras in a sequence, we set $c_{min} = 0.05 \, c_M$ and $c_{max} = 0.2 \, c_M$, and $t_{min} = -8$ and $t_{max} = 8$.

\noindent\textbf{Camera viewpoint angle.} Conditioning views must have a viewing direction with cosine similarity within $\alpha_{min} < cos^{-1} \frac{\textbf{v}_c^n \cdot \textbf{v}_t}{||\textbf{v}_c^n||||\textbf{v}_t||} < \alpha_{max}$, where $\textbf{v} = \textbf{R}^{-1} \times [0,0,1]^T$ is a vector pointing forward (positive $z$) relative to a world-to-camera rotation matrix $\textbf{R}$. In all experiments, we set $\alpha_{min} = 0$ and $\alpha_{max}=\pi/2$ for depth generation, to avoid supervision from sparse reconstructions, and $\alpha_{min} = 0$ and $\alpha_{max}=\pi$ for image generation, to promote extrapolation to novel viewpoints.

\noindent\textbf{Pointcloud overlapping.} Whenever depth maps are available, we set a threshold $p_{min}$ on the percentage of how many valid pixels of each conditioning view are projected onto the target view. For each pixel $\textbf{p}_c^n=\{u,v\}$ with depth $d$ from a conditioning view $\textbf{I}_c^n$ with depth $\textbf{D}_c^n$, we can obtain its projection $\textbf{p}_t'$ and depth $d'_t$ onto the target view via:
\begin{equation}
\begin{bmatrix}
u' \\ v' \\ d' \\ 1
\end{bmatrix}_t
=
\Big( \tilde{\textbf{K}}_c^n
\textbf{T}_c^n
\Big)^{-1}
\begin{bmatrix}
u \\ v \\ d \\ 1
\end{bmatrix}_c^n
\Big(
\textbf{T}_t
\tilde{\textbf{K}}_t 
\Big)
\end{equation} 
where $\scriptsize \tilde{\textbf{K}} = 
\begin{bmatrix} 
\textbf{K} & \textbf{0} \\
\textbf{0} & 1 
\end{bmatrix}
\in \mathbb{R}^{4 \times 4}$ is a homogeneous intrinsics matrix. 
A projected point is considered valid if $u' \in [0,H]$ and $v' \in [0,W]$, i.e. if its projected coordinates are inside the target view. 
We set $p_{min} = 30\%$, and additionally discard samples with less than $64$ valid projected pixels.

\begin{figure*}[t!]
\begin{center}
    \centering
\vspace{4mm}
\includegraphics[width=0.32\textwidth]{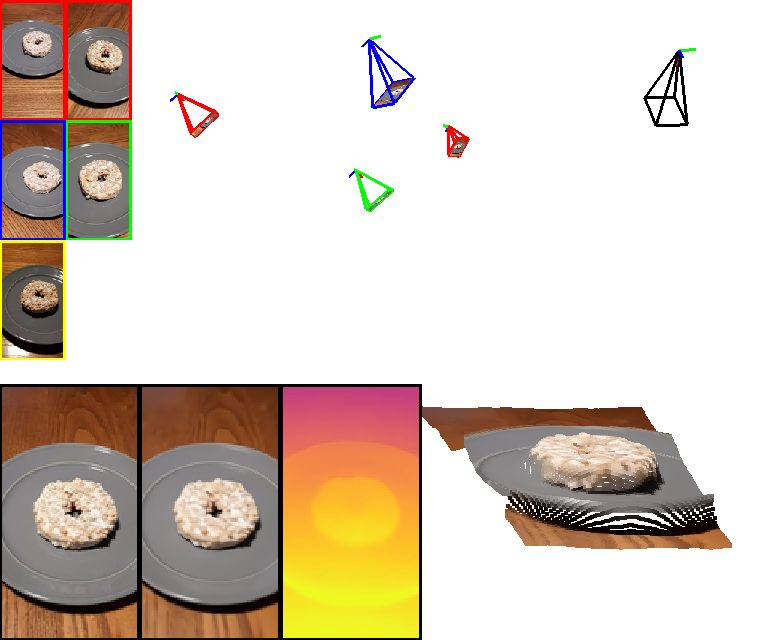}
\includegraphics[width=0.32\textwidth]{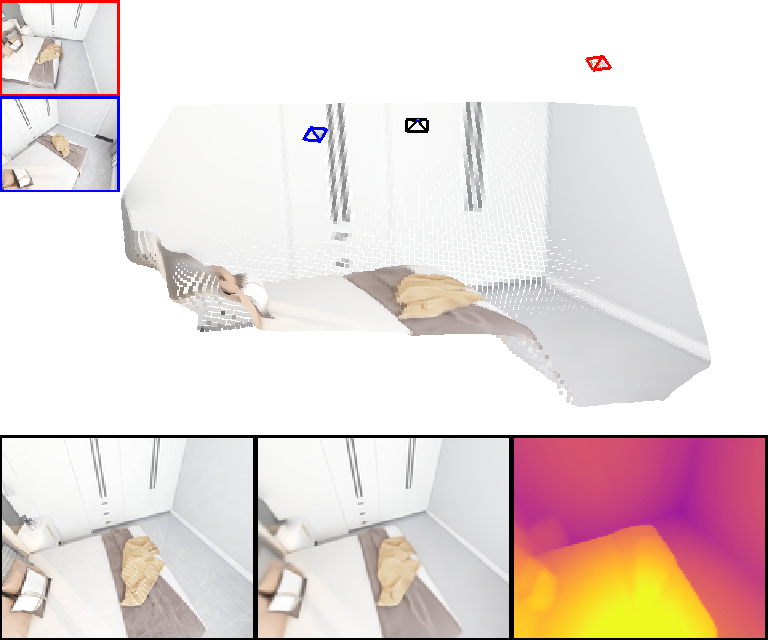}
\includegraphics[width=0.32\textwidth]{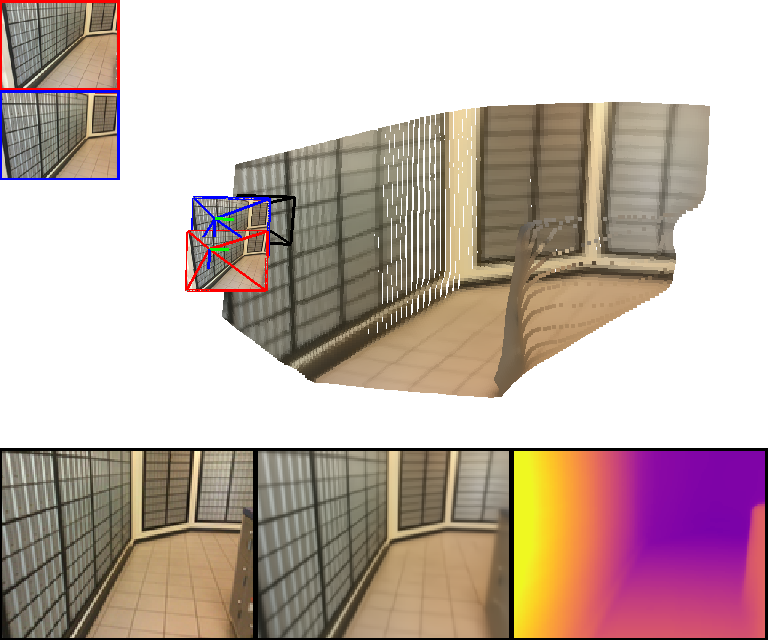}
\vspace{4mm} \\
\includegraphics[width=0.32\textwidth]{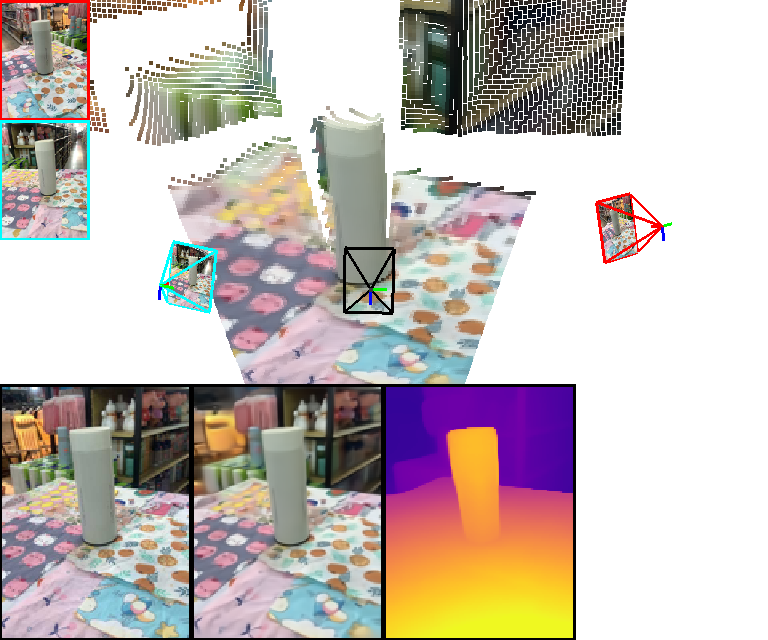}
\includegraphics[width=0.32\textwidth]{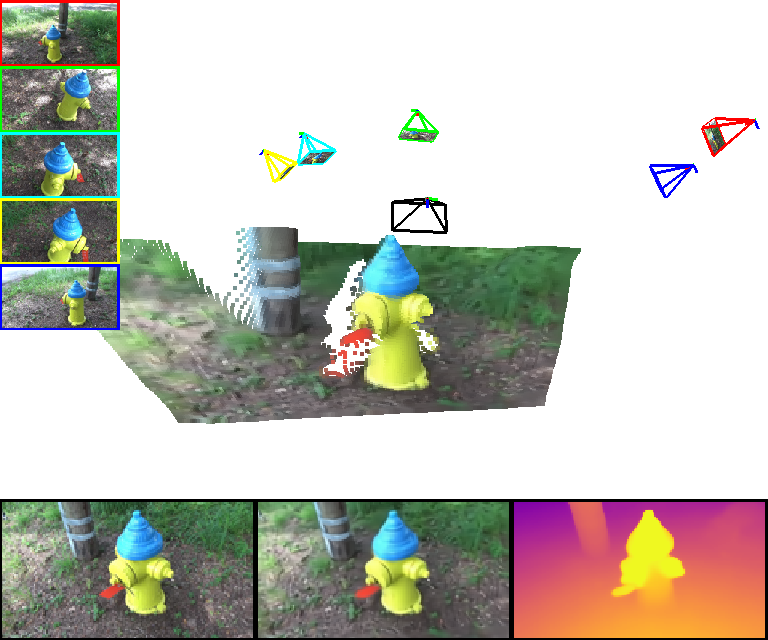}
\includegraphics[width=0.32\textwidth]{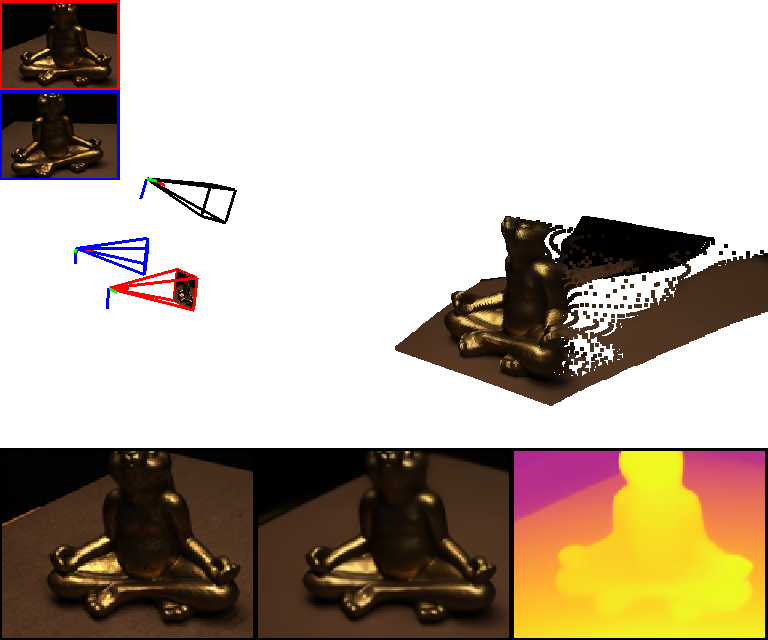}
\vspace{4mm} \\
\includegraphics[width=0.32\textwidth]{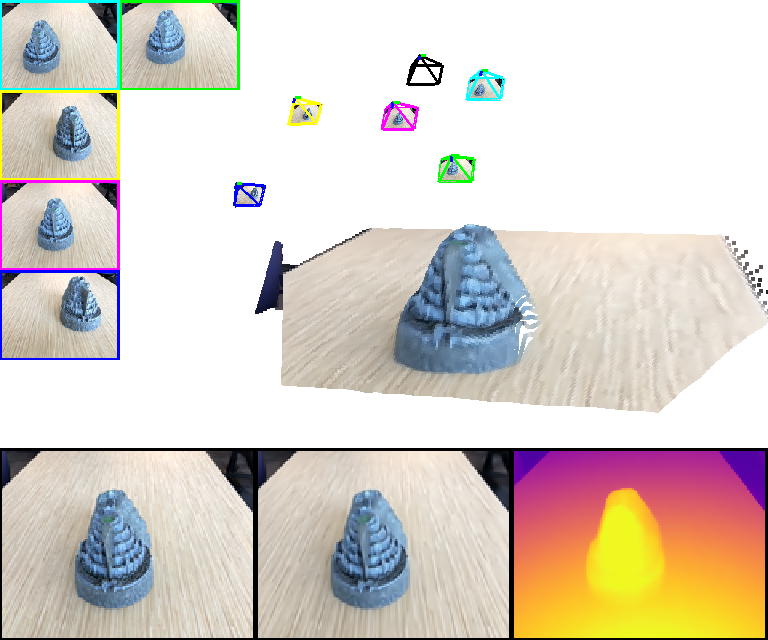}
\includegraphics[width=0.32\textwidth]{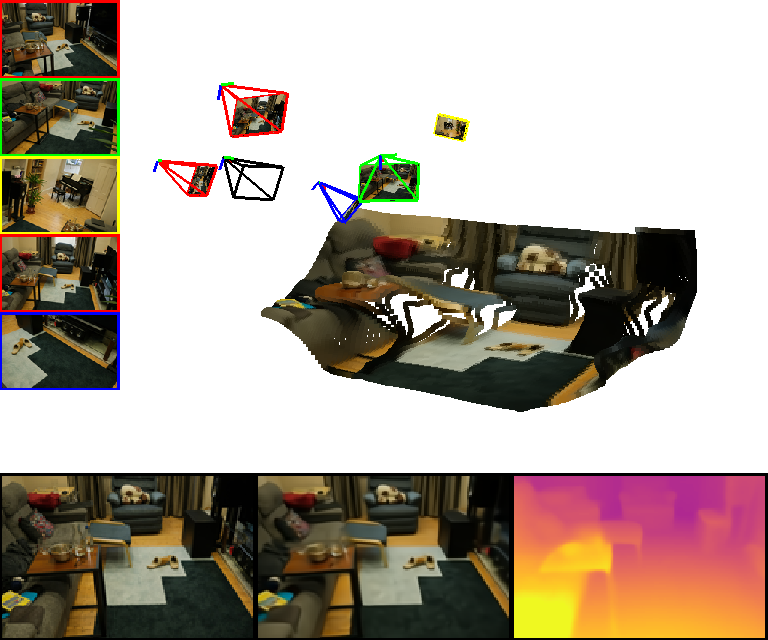}
\includegraphics[width=0.32\textwidth]{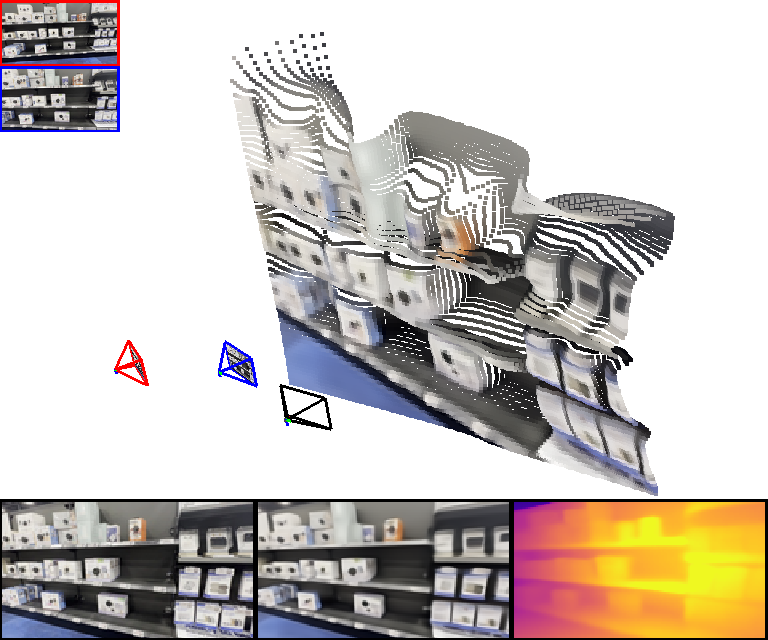}
\vspace{4mm} \\
\includegraphics[width=0.32\textwidth]{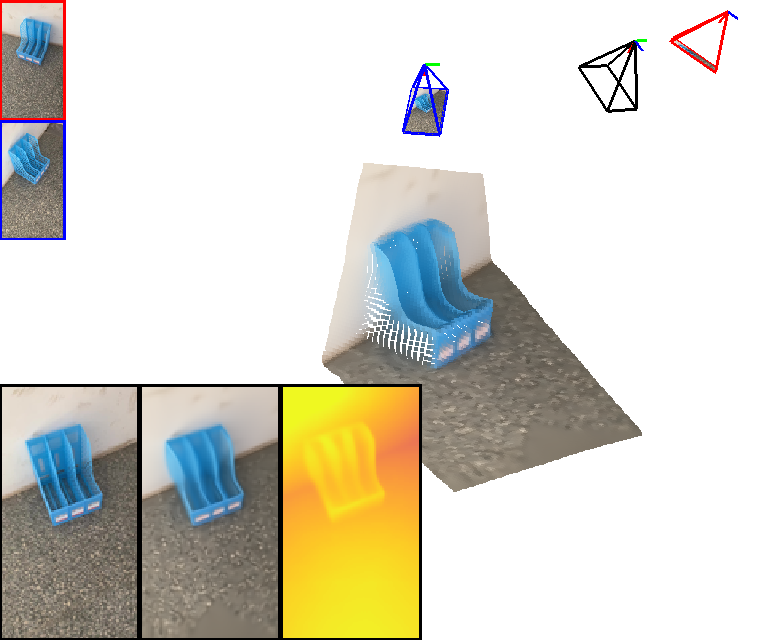}
\includegraphics[width=0.32\textwidth]{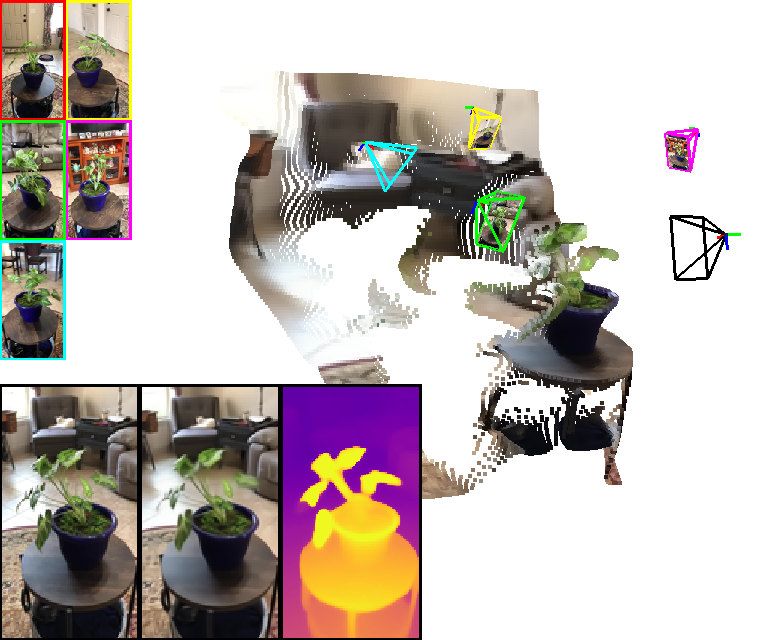}
\includegraphics[width=0.32\textwidth]{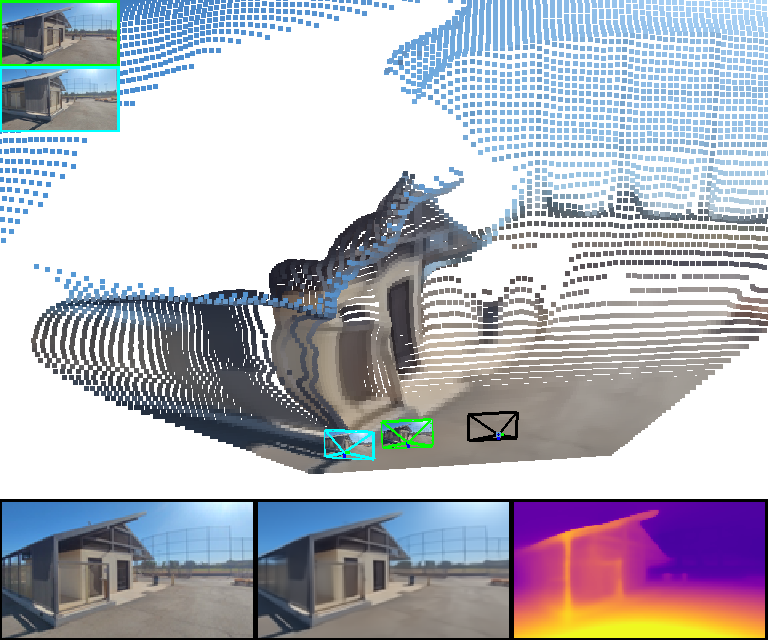}
\caption{
    \textbf{Zero-Shot \Acronym novel view and depth synthesis results} randomly sampled from different evaluation benchmarks and in-the-wild datasets. Top left images are conditioning views (colored cameras), and bottom images are the target view (black camera), showing from left-to-right: ground-truth image, predicted image, and predicted depth map. These predictions are used to produce a colored 3D pointcloud observed from the target viewpoint. 
}
\label{fig:supp_qualitative_single}
\end{center}
\vspace{-4mm}
\end{figure*}

\begin{figure*}[t!]
\renewcommand{\arraystretch}{0.9}
\centering
\scriptsize
\setlength{\tabcolsep}{0.8em}
\center
\includegraphics[width=0.4\textwidth,height=5cm]{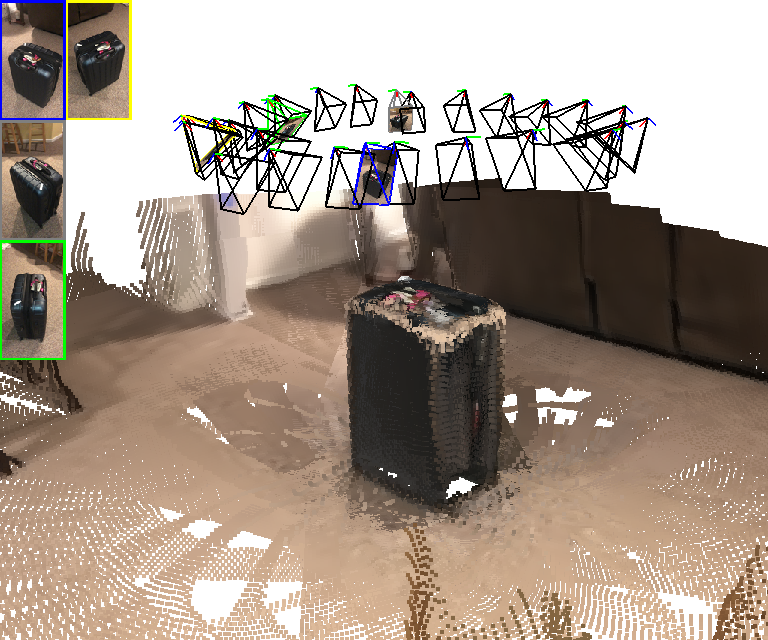}
\hspace{5mm}
\includegraphics[width=0.4\textwidth,height=5cm]{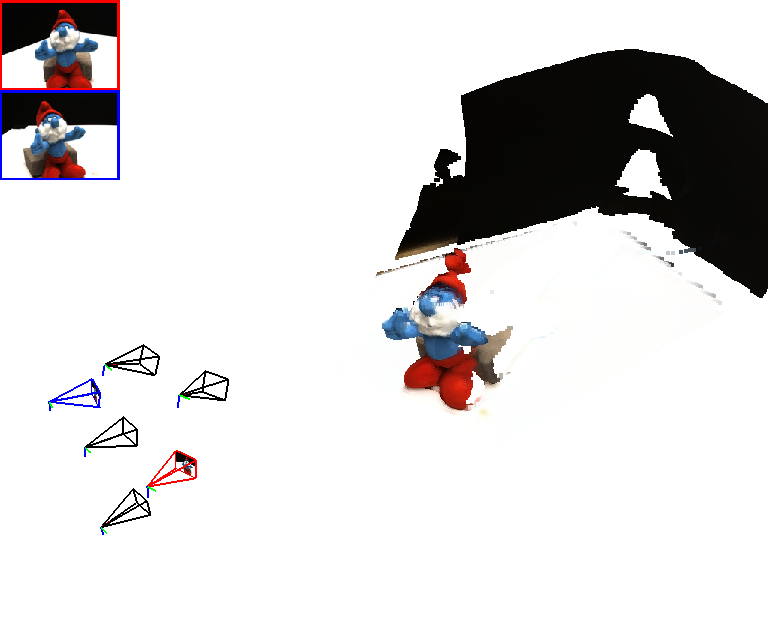}
\\
\includegraphics[width=0.42\textwidth,height=5cm]{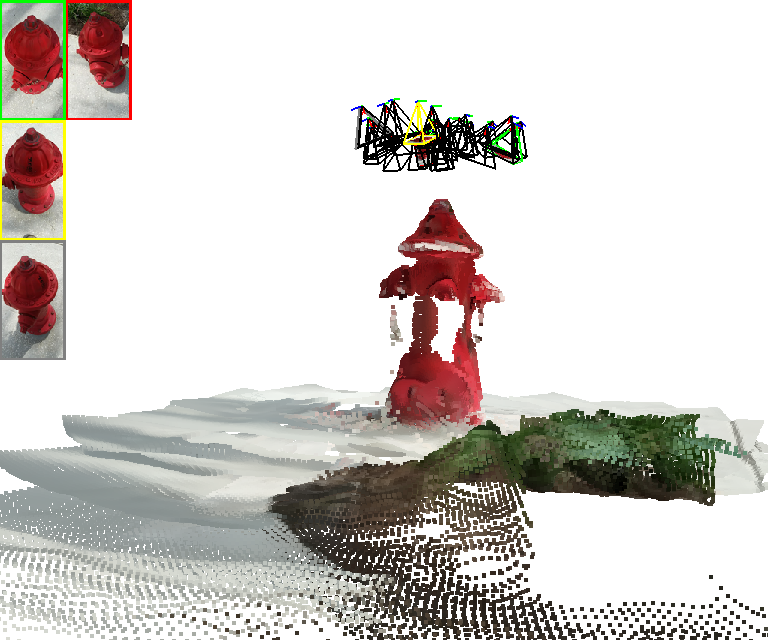}
\hspace{5mm}
\includegraphics[width=0.42\textwidth,height=5cm]{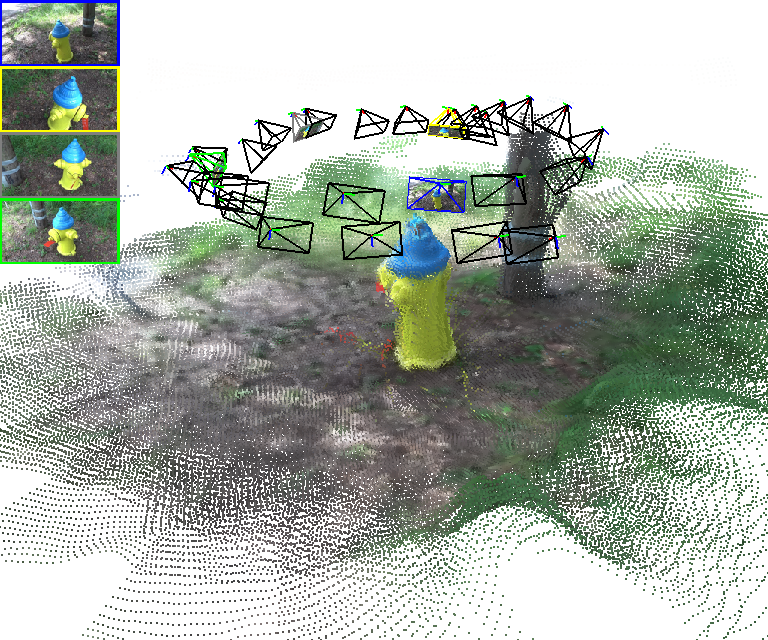}
\\
\includegraphics[width=0.42\textwidth,height=5cm]{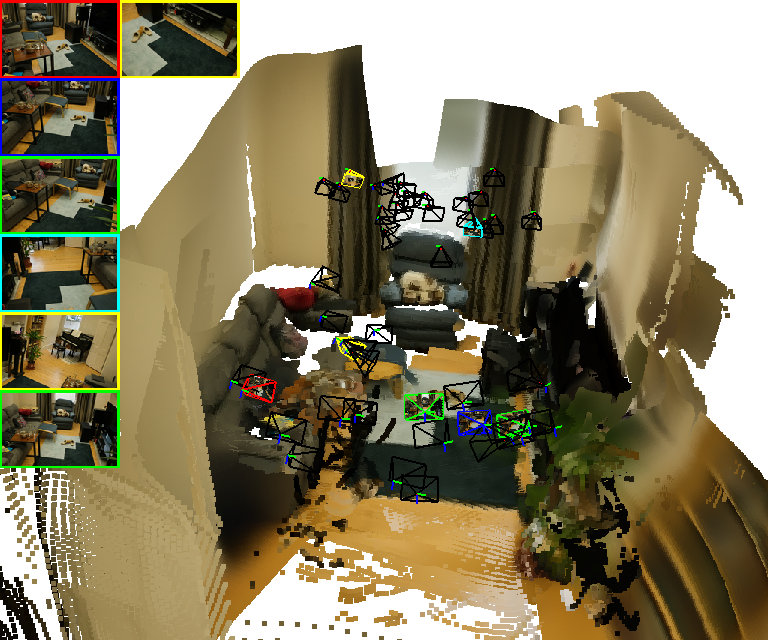}
\hspace{5mm}
\includegraphics[width=0.42\textwidth,height=5cm]{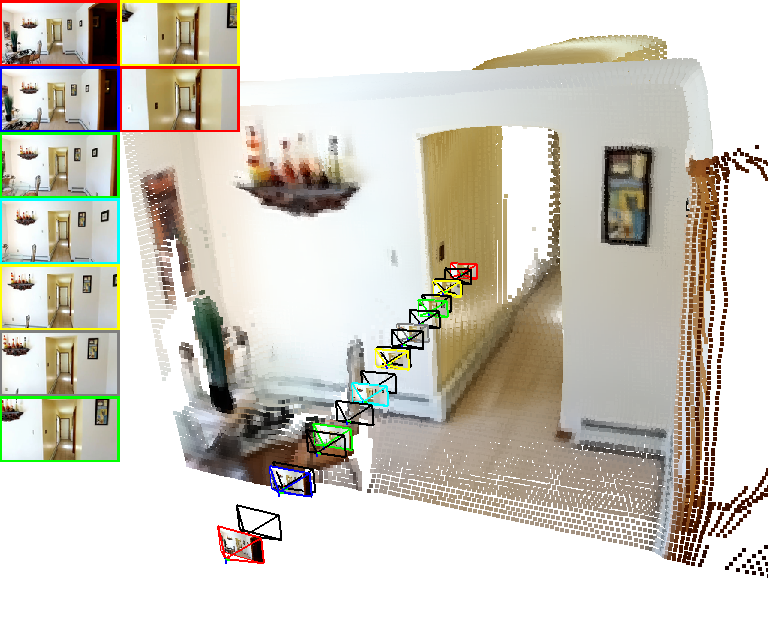}
\\
\includegraphics[width=0.42\textwidth,height=5cm]{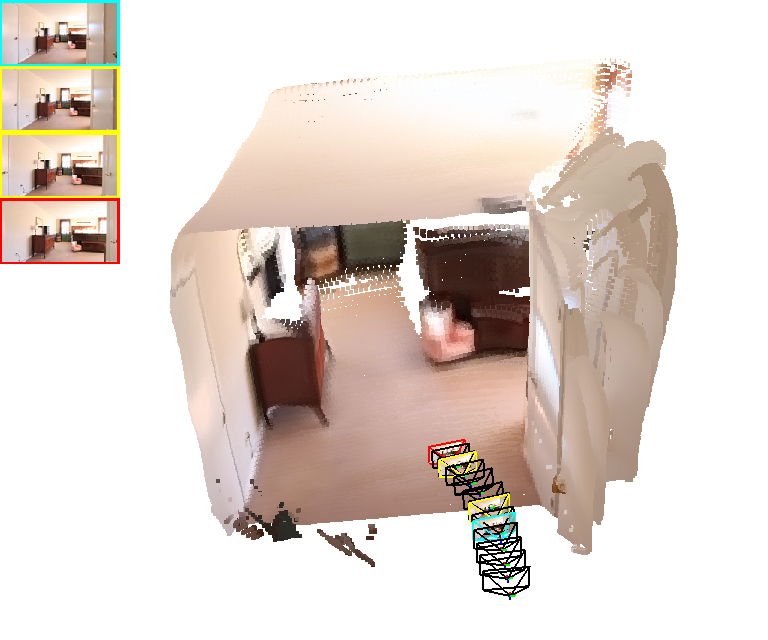}
\hspace{5mm}
\includegraphics[width=0.42\textwidth,height=5cm]{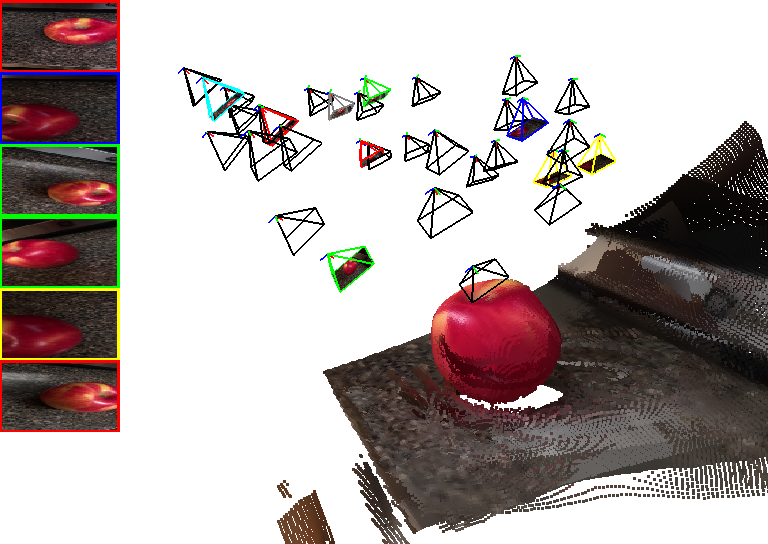}

\caption{\textbf{Accumulated \Acronym pointclouds}, obtained by generating novel images and depth maps from various viewpoints (black cameras), using the same conditioning views (colored cameras), and stacking them together without any post-processing. Our zero-shot architecture is capable of directly generating multi-view consistent predictions that match the scale from conditioning cameras.
}
\label{fig:surround}
\end{figure*}

These criteria were used as a pre-processing step, to generate a list of valid training samples. 
We will open-source dataloaders for all training and validation datasets, to facilitate the reproduction of our work, as well as the list of valid samples used in our experiments. 
We use \texttt{Webdataset}~\cite{webdataset} to optimize storage and training efficiency.

\begin{figure*}[t!]
\renewcommand{\arraystretch}{0.9}
\centering
\scriptsize
\setlength{\tabcolsep}{0.8em}
\center
\begin{tabular}{cc}
\includegraphics[width=0.3\textwidth,height=4.1cm]{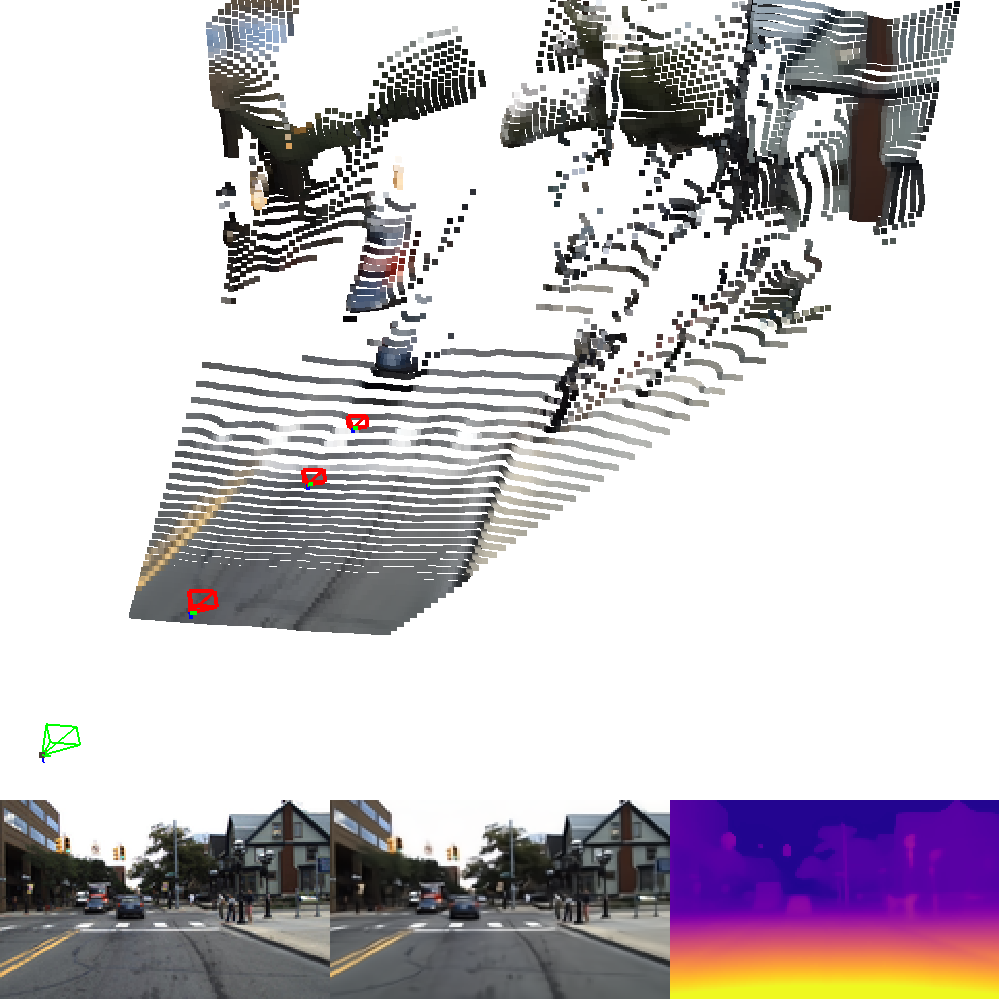}
\includegraphics[width=0.3\textwidth,height=4.1cm]{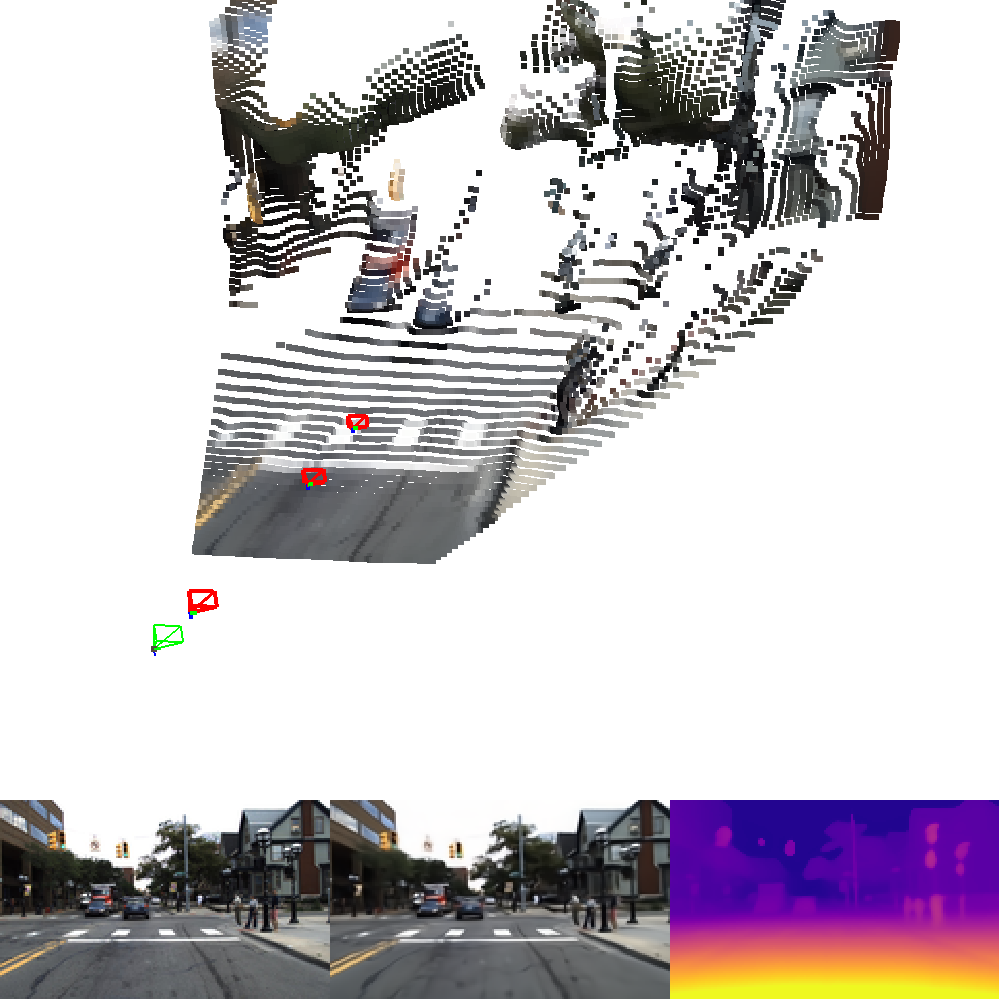}
&
\hspace{-8mm}
\multirow{2}{*}[21.5mm]{  
\hspace{3mm}
\subfloat{
\vspace{-9.5mm}
  \includegraphics[width=0.35\textwidth]{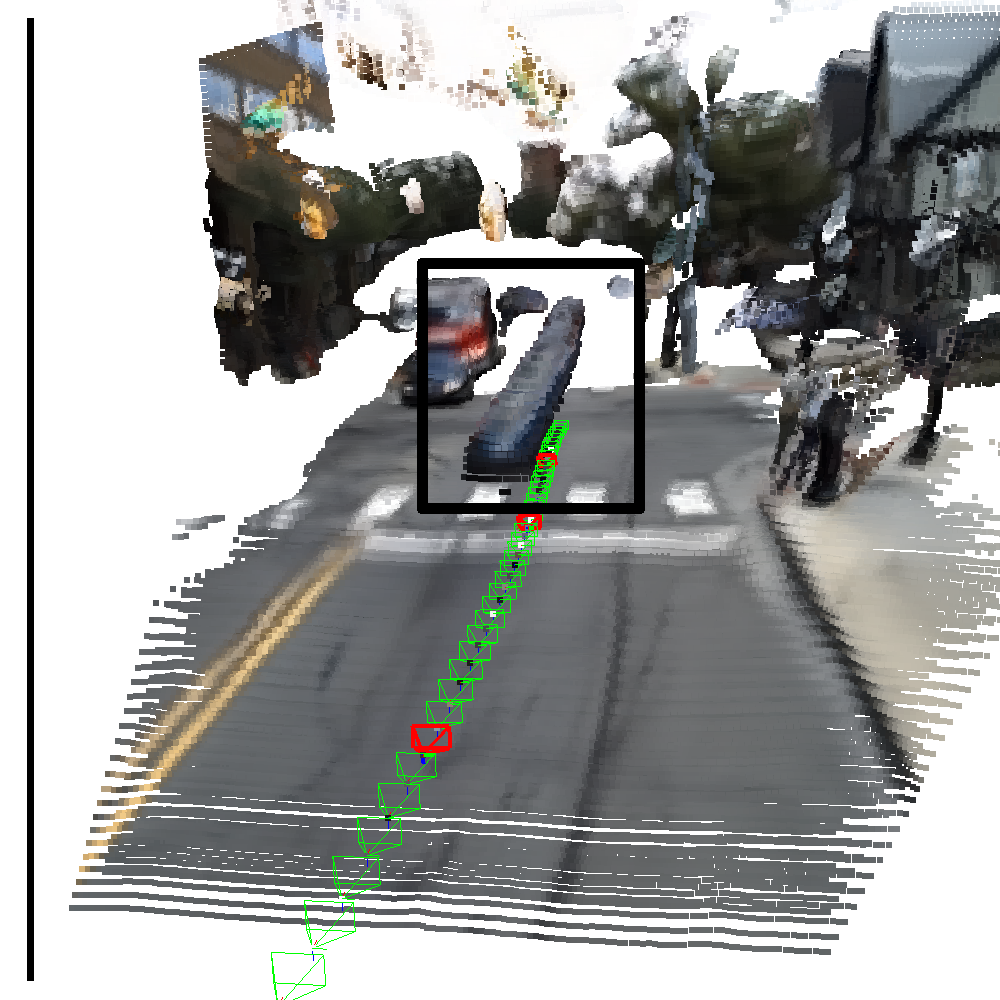}
}}
\vspace{8mm}
\\
\includegraphics[width=0.3\textwidth,height=4.1cm]{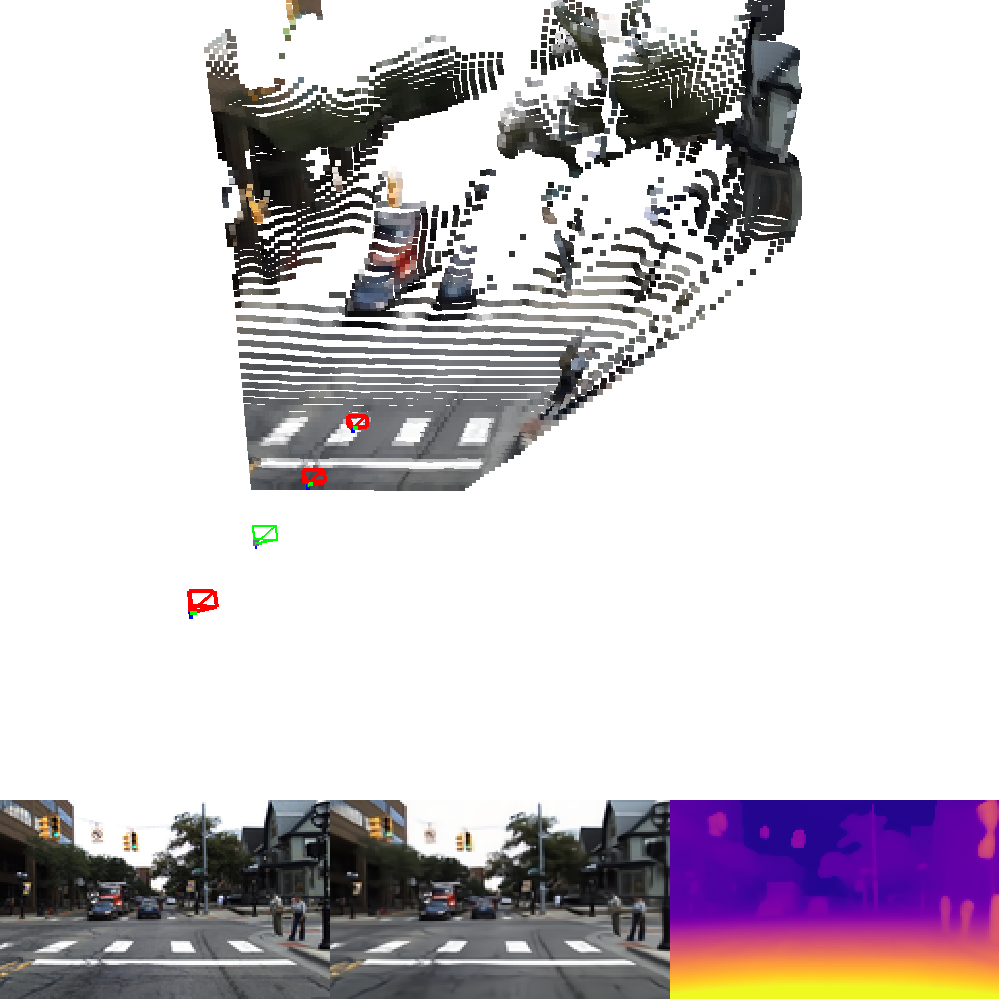}
\includegraphics[width=0.3\textwidth,height=4.1cm]{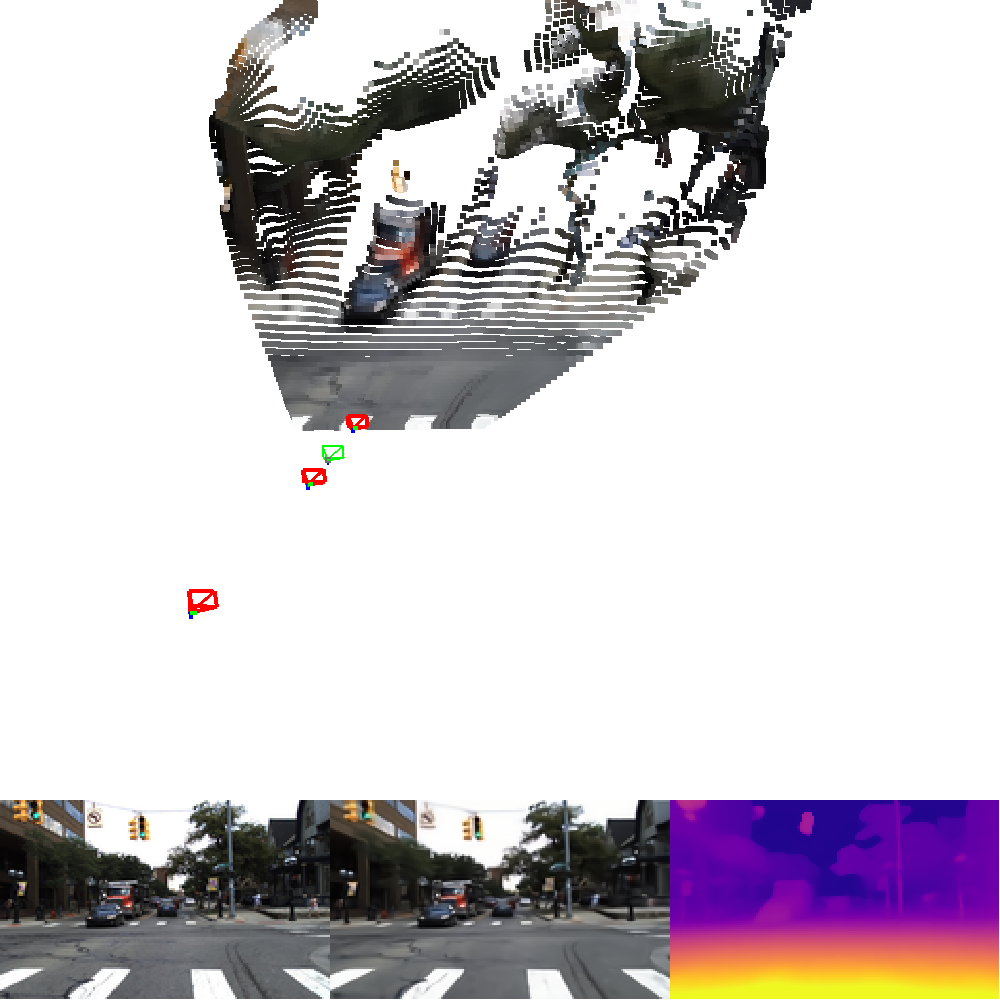}
\\
\end{tabular}
\begin{tabular}{cc}
\includegraphics[width=0.3\textwidth,height=4.1cm]{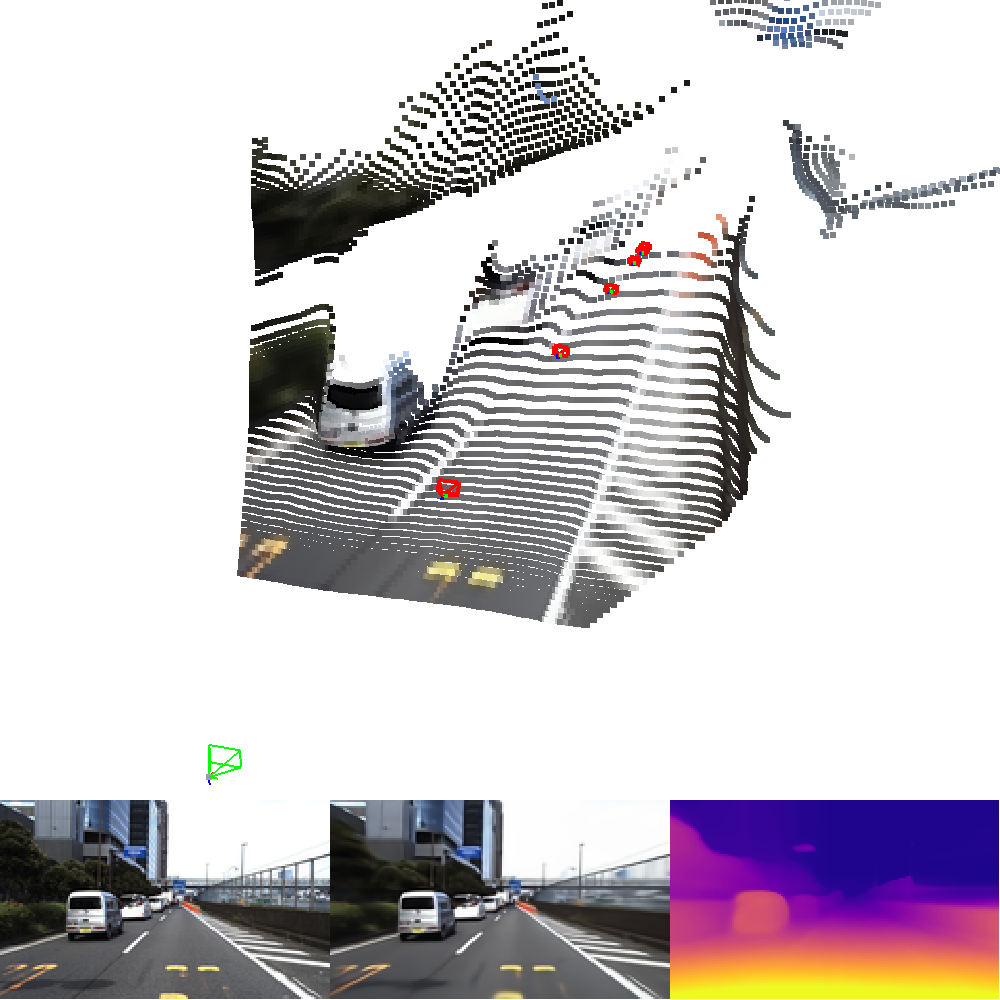}
\includegraphics[width=0.3\textwidth,height=4.1cm]{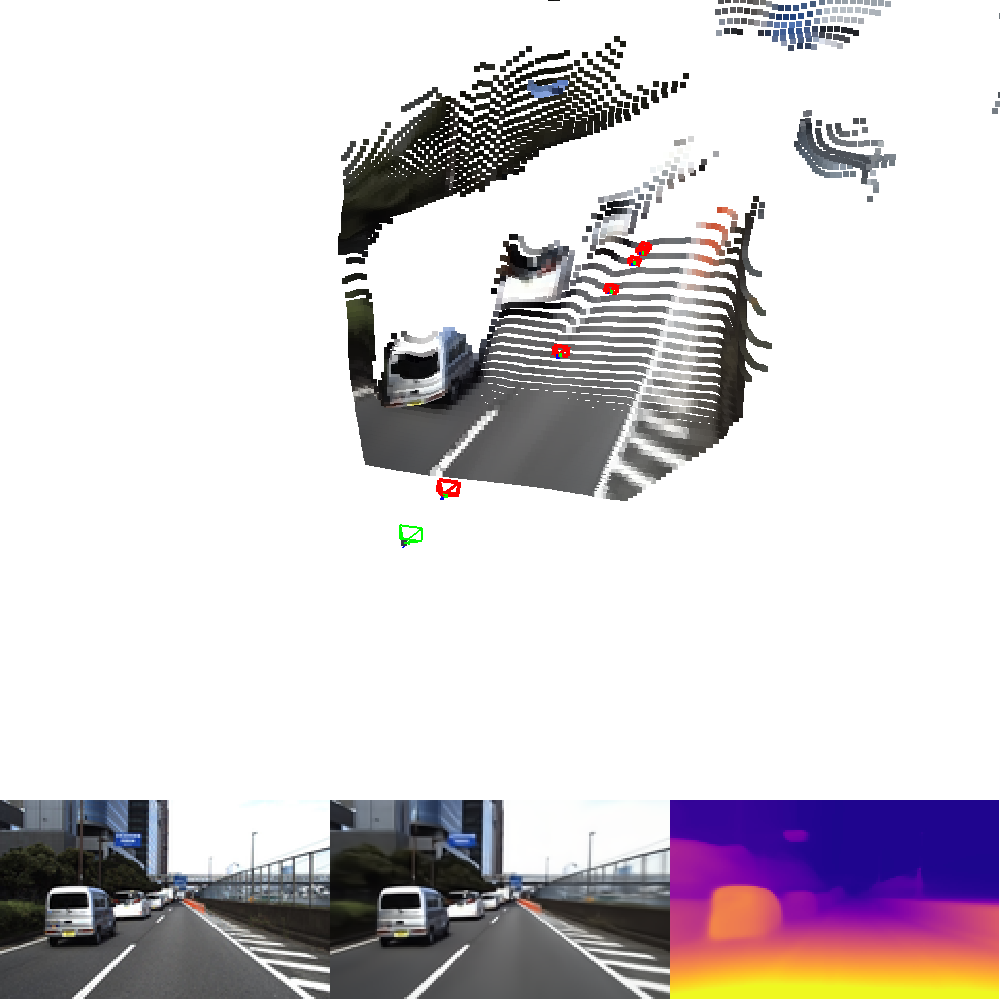}
&
\hspace{-8mm}
\multirow{2}{*}[21.5mm]{  
\hspace{3mm}
\subfloat{
\vspace{-9.5mm}
  \includegraphics[width=0.35\textwidth]{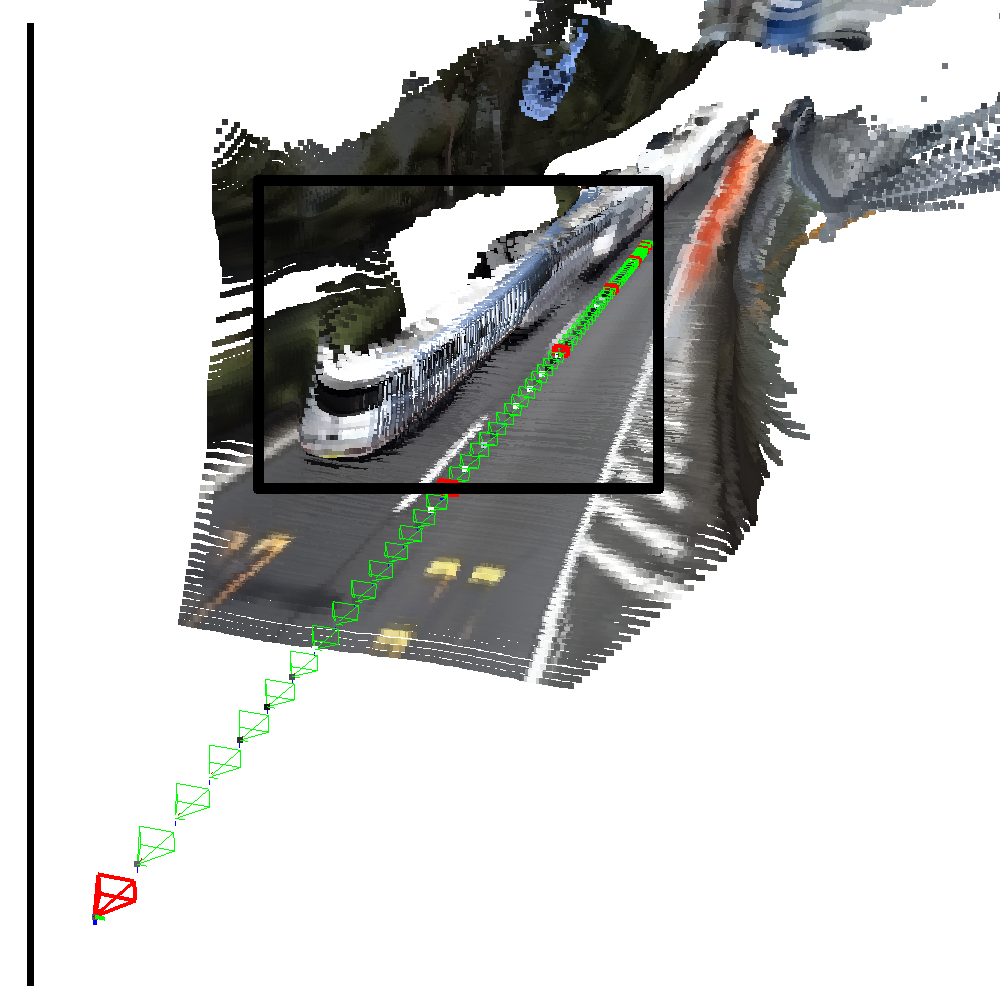}
}}
\\
\includegraphics[width=0.3\textwidth,height=4.1cm]{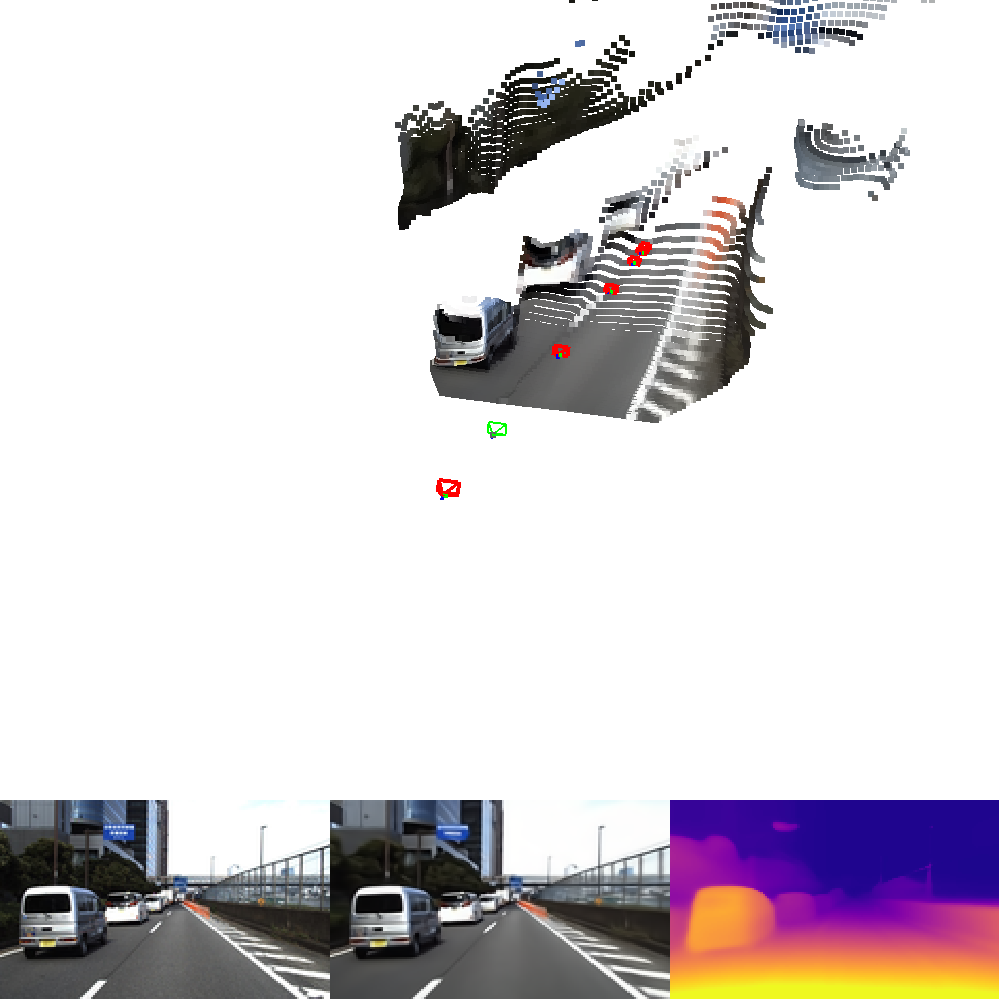}
\includegraphics[width=0.3\textwidth,height=4.1cm]{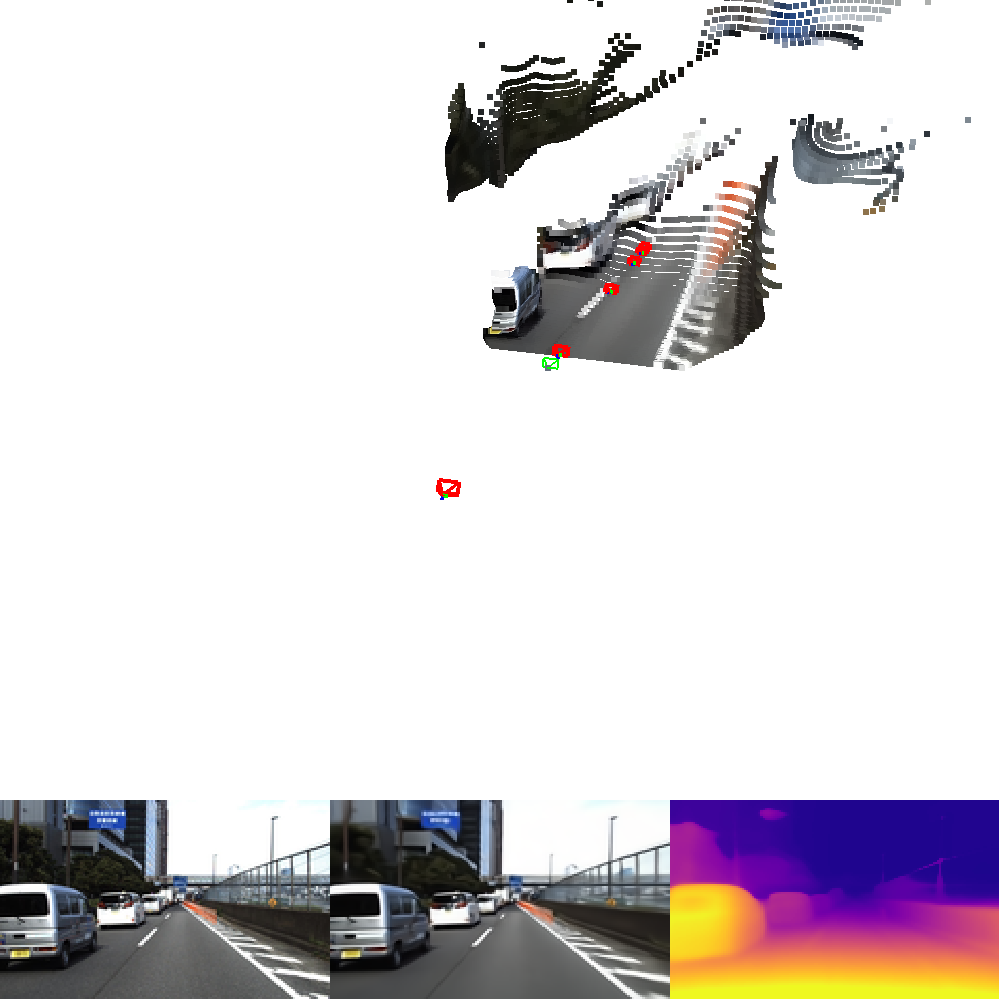}
\\
\end{tabular}
\caption{\textbf{Accumulated MVGD pointclouds on a dynamic dataset~\cite{packnet}}. (left) Red cameras are used as conditioning views, and novel images and depth maps are generated from green cameras. (right) Colored pointclouds are calculated from these predictions and stacked together without any post-processing. Even though MVGD does not explicitly model dynamic objects, it implicitly learns how the scene should change when interpolating between views with objects in different locations (e.g., moving cars), while keeping the remainder static.}
\label{fig:motion}
\vspace{-3mm}
\end{figure*}

\begin{figure*}[t!]
\begin{center}
\centering
\includegraphics[trim={2.0cm 3.0cm 0 0},clip,width=0.99\textwidth,height=6.6cm]{
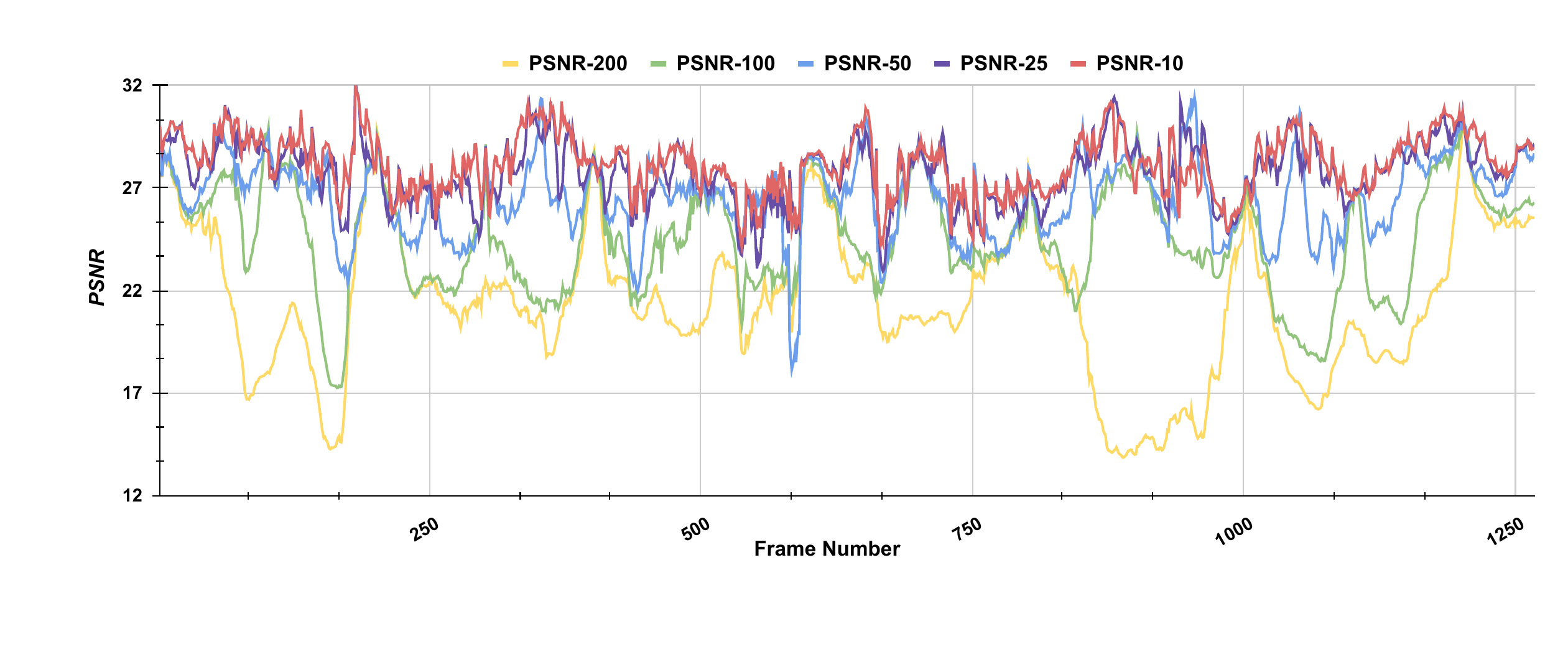}
\\
\includegraphics[trim={2cm 3.0cm 0 0},clip,width=0.99\textwidth,height=6.6cm]{
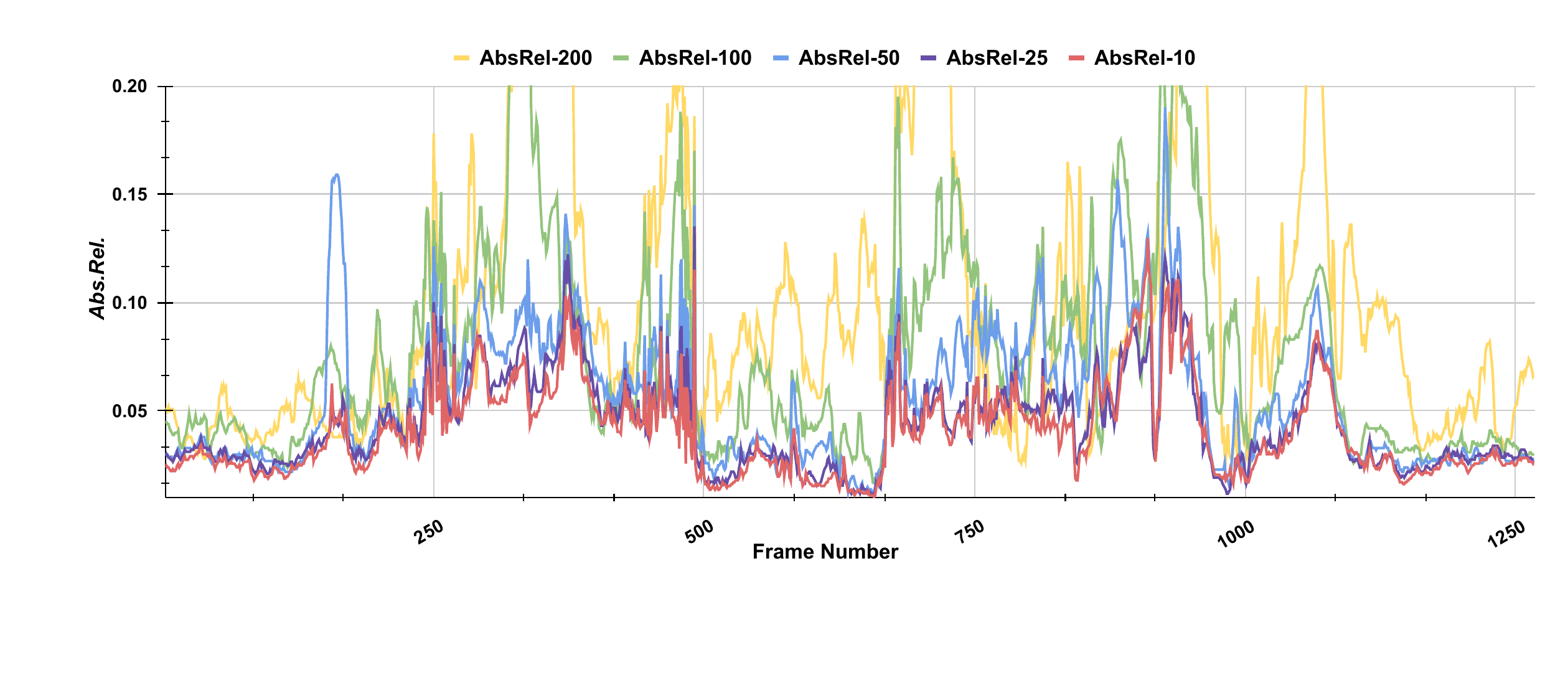}
\\
\caption{
    \textbf{\Acronym per-frame novel view and depth synthesis results} using different numbers of fixed conditioning views, on ScanNet (scene $0086\_02$, with $1267$ images).  The same model from all experiments reported in the main paper was used. Legend numbers indicate the stride (i.e., how many target images are positioned between each conditioning view). As expected, results consistently improve as more input information is available, eventually plateauing at around $100$ conditioning views.
}
\label{fig:interpolation}
\end{center}
\vspace{-5mm}
\end{figure*}

\begin{figure*}[t!]
\begin{center}
\centering
\includegraphics[trim={2cm 3.0cm 0 0},clip,width=0.95\textwidth]{
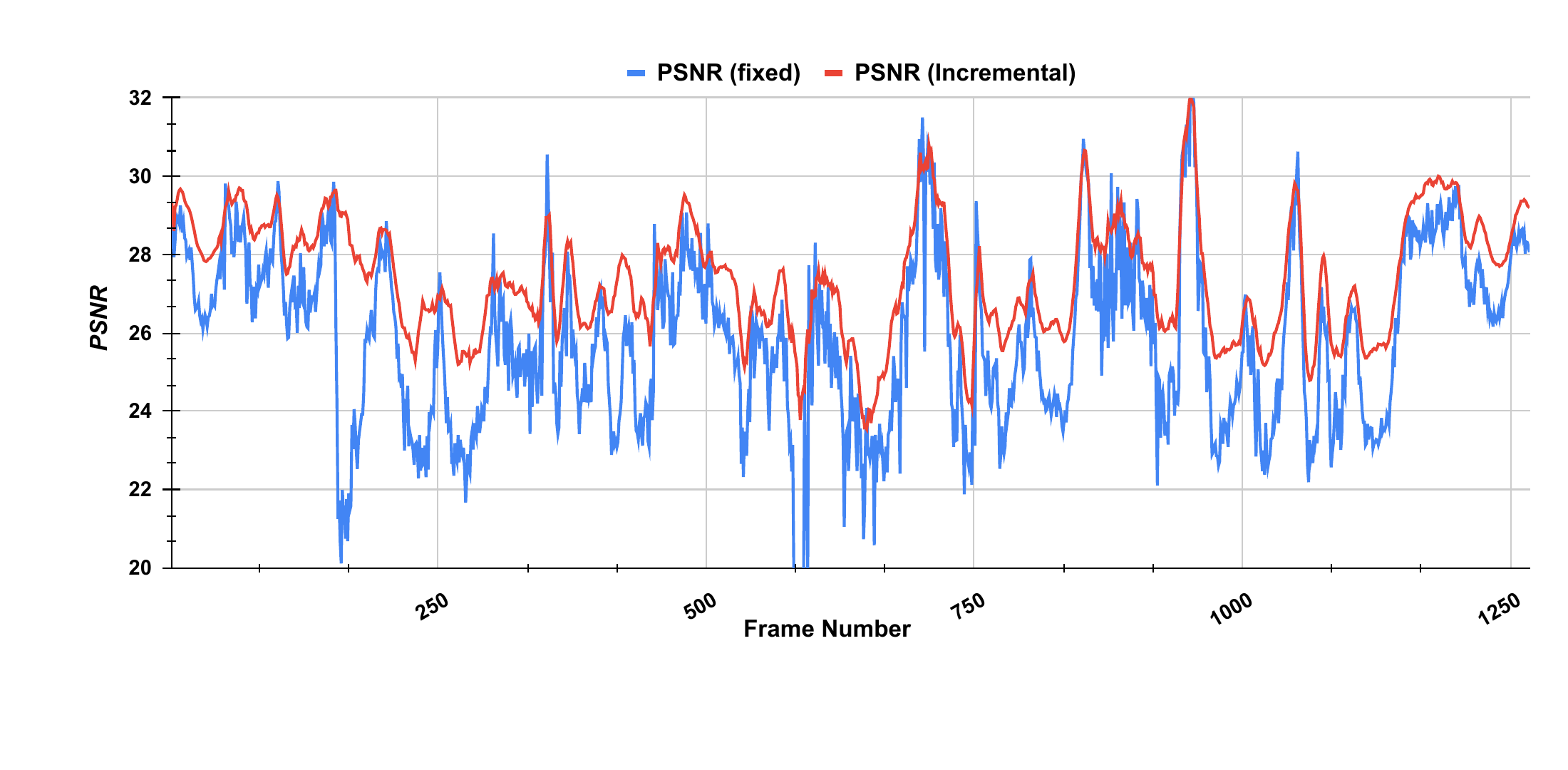}
\vspace{-5mm}
\includegraphics[trim={2cm 2.0cm 0 0.5cm},clip,width=0.95\textwidth]{
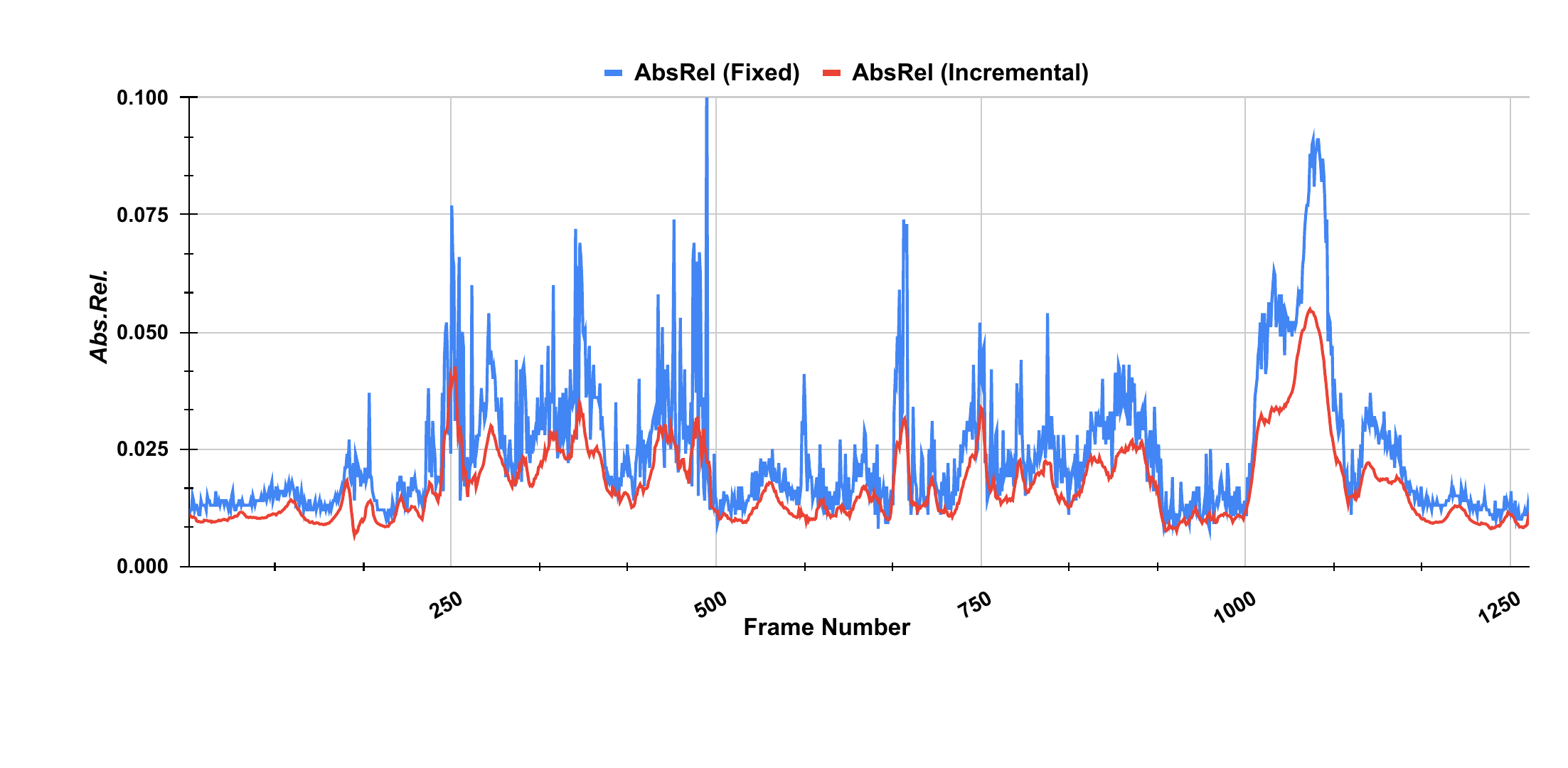}
\\
\caption{
    \textbf{\Acronym per-frame novel view and depth synthesis results} with and without our proposed incremental conditioning strategy, on ScanNet (scene $0086\_02$, with $1267$ images). The same model from all experiments reported in the main paper was used. The blue line indicates a fixed number (25) of conditioning views, evenly spaced with a stride of $50$. The red line indicates our proposed incremental conditioning strategy, in which each new generation uses previously generated images as additional conditioning. This strategy leads to consistently better and more stable results in novel view and depth synthesis, especially in regions further away from conditioning views.  
}
\label{fig:incremental}
\end{center}
\vspace{-5mm}
\end{figure*}

\begin{figure*}[t!]
\begin{center}
\centering
\subfloat[\textbf{149} frames, \textbf{5} initial conditioning views.]{
\includegraphics[width=0.48\textwidth,height=5.3cm]{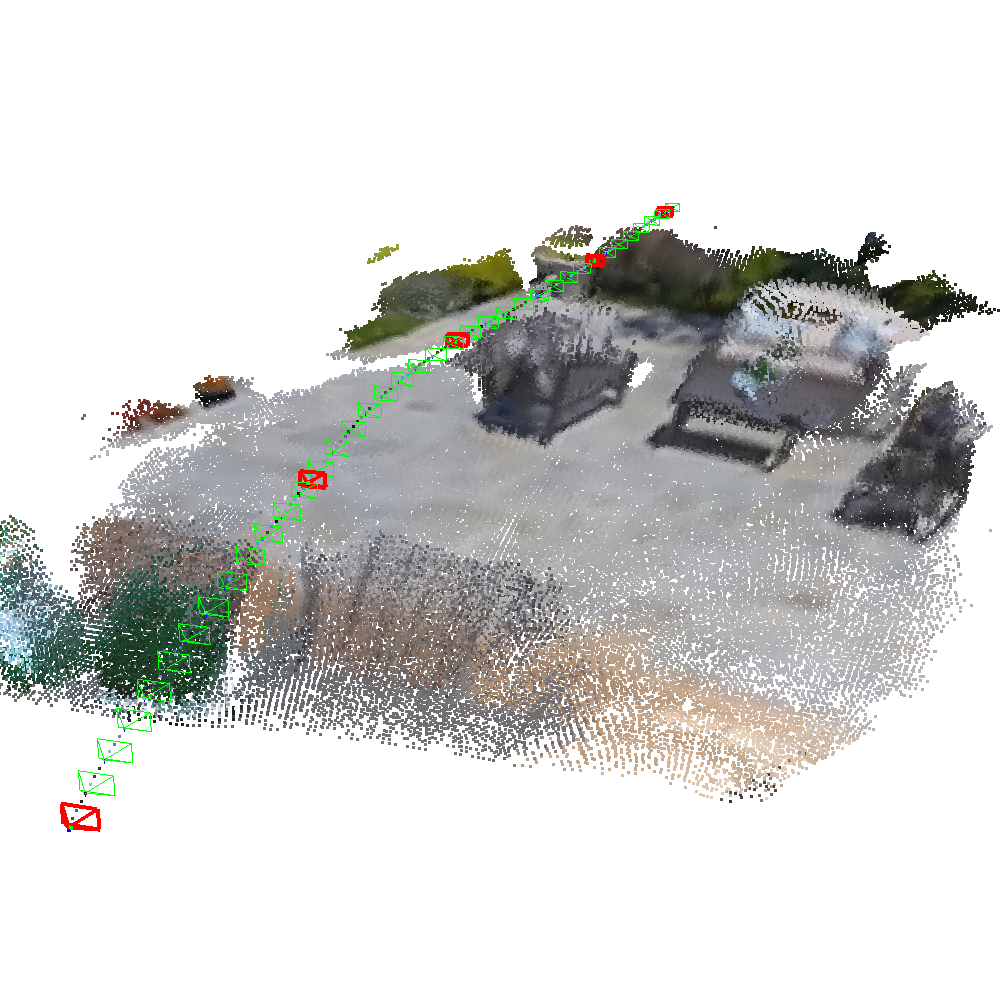}}
\subfloat[\textbf{202} frames, \textbf{7} initial conditioning views.]{
\includegraphics[width=0.48\textwidth,height=5.3cm]{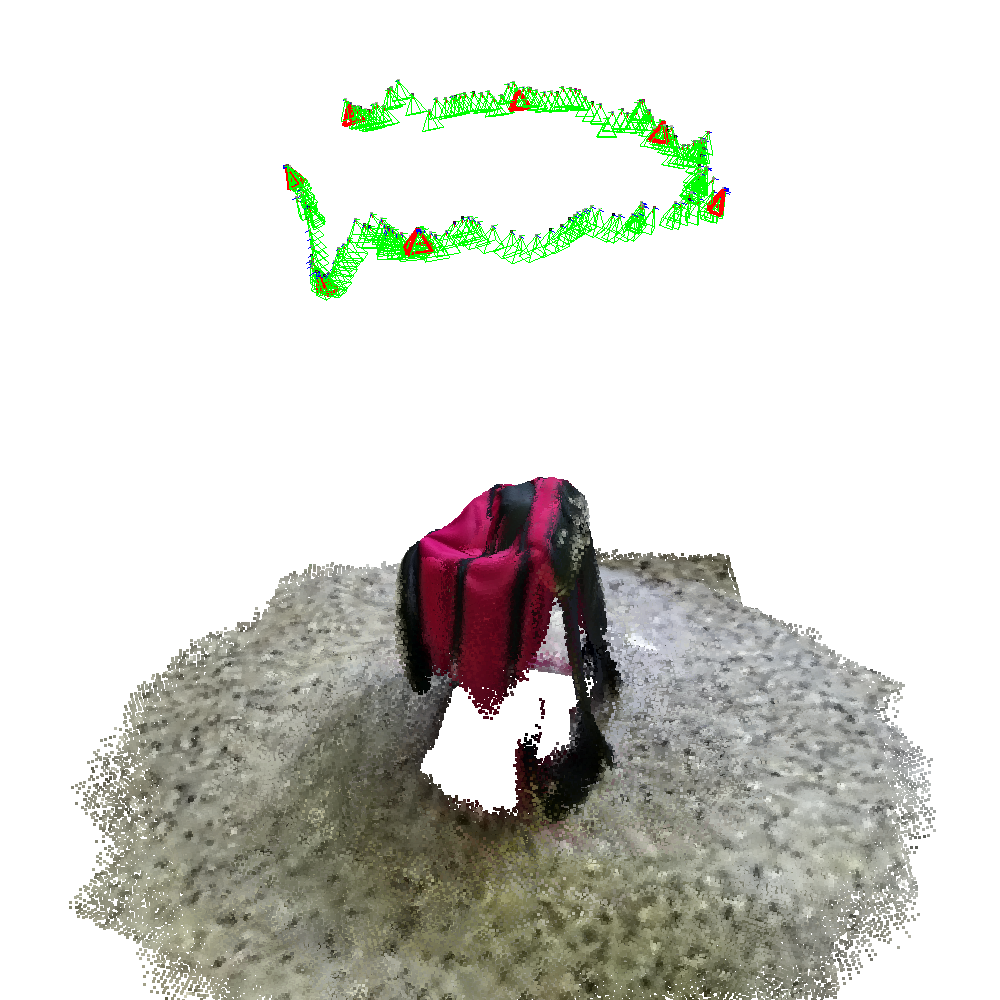}}
\vspace{2.3mm} 
\\
\subfloat[\textbf{252} frames, \textbf{5} initial conditioning views.]{
\includegraphics[width=0.48\textwidth,height=5.3cm]{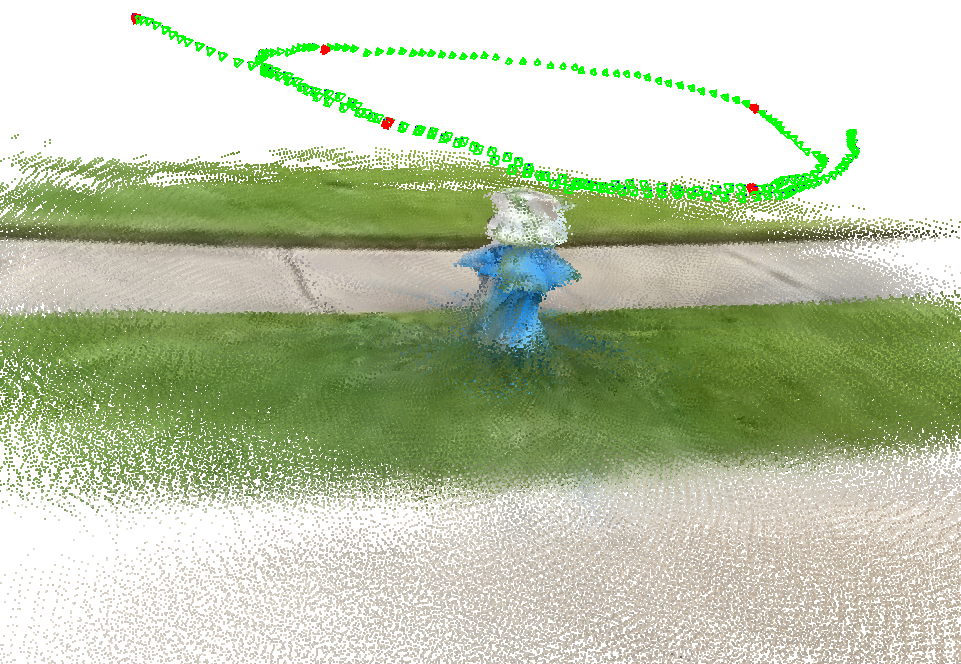}}
\subfloat[\textbf{337} frames, \textbf{6} initial conditioning views.]{
\includegraphics[width=0.48\textwidth,height=5.3cm]{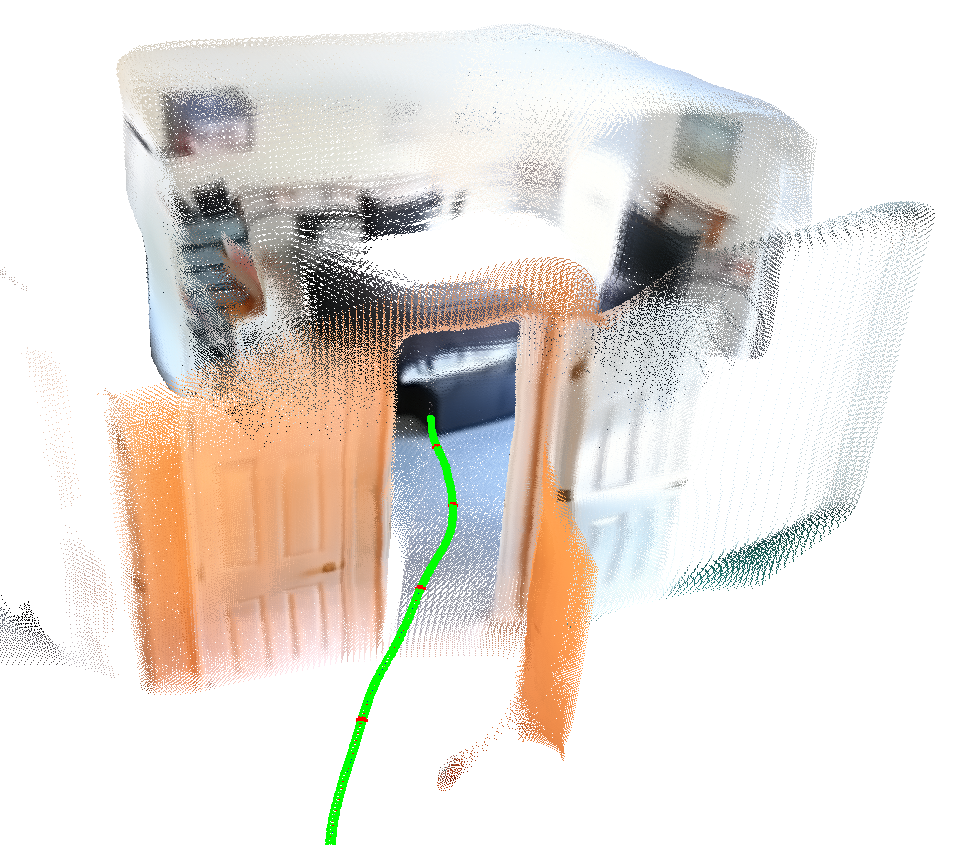}}
\vspace{2mm}
 \\
\subfloat[\textbf{1268} frames, \textbf{25} initial conditioning views.]{
\includegraphics[width=0.48\textwidth,height=5.3cm]{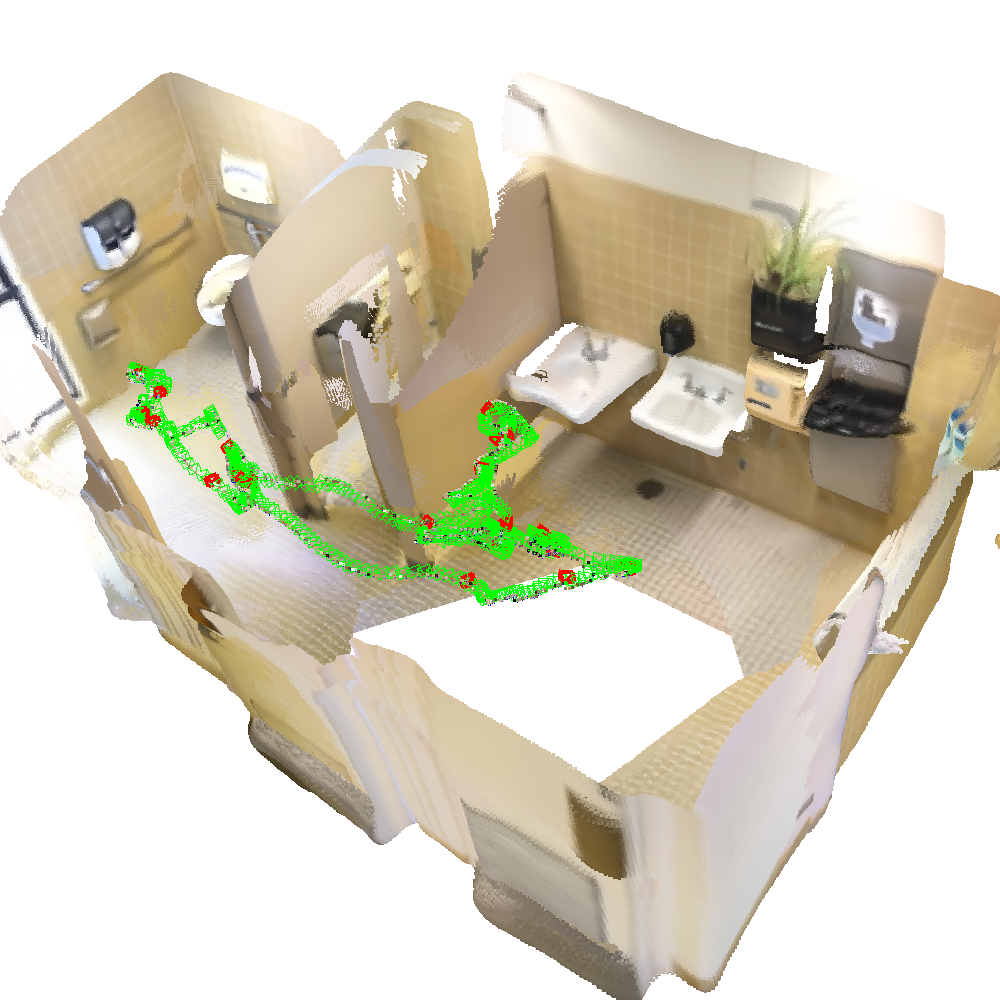}}
\subfloat[\textbf{5578} frames, \textbf{60} initial conditioning views.]{
\includegraphics[width=0.48\textwidth,height=5.3cm]{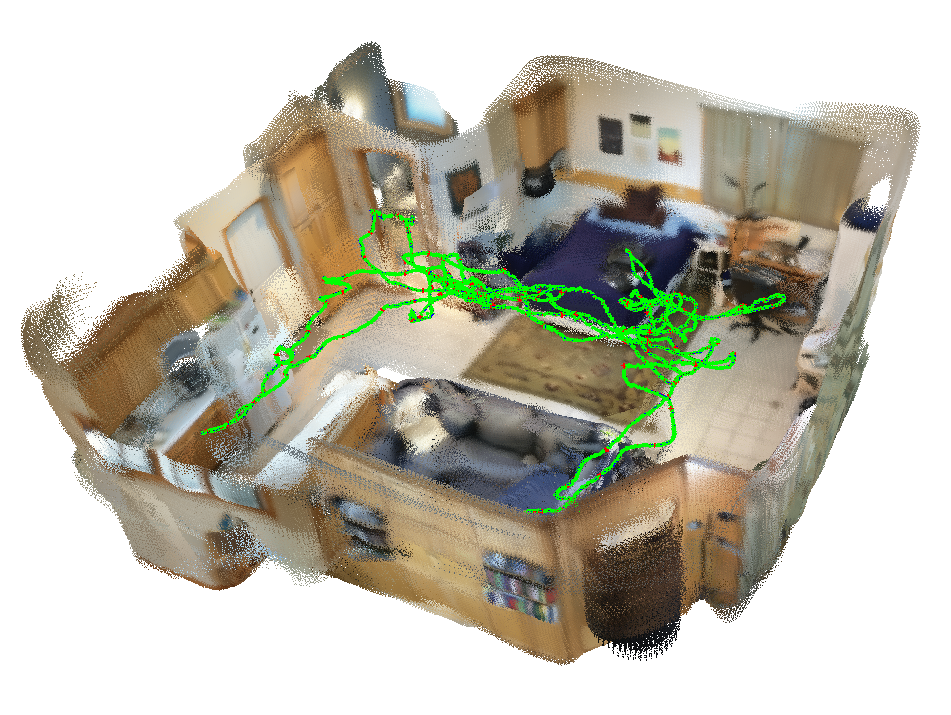}}
\caption{
    \textbf{\Acronym novel view and depth synthesis results} using our proposed incremental conditioning strategy. Red cameras indicate initial conditioning views, used to generate predictions for green cameras (ordered from closest to furthest away from the initial conditioning views). After each generation, the predicted image is added to the set of conditioning views for future generations. Even though \Acronym was trained using only $2 - 5$ conditioning views, it can directly scale to thousands on a single GPU. 
}
\label{fig:supp_qualitative_all}
\end{center}
\vspace{-6mm}
\end{figure*}

\section{Additional Qualitative Examples}

In Figure \ref{fig:supp_qualitative_single} we provide additional qualitative results of \Acronym in different evaluation benchmarks, as well as in-the-wild images from different sources (complementing Figure 3 of the main paper). 
Conditioning images are shown in the top left, with corresponding cameras (denoted by different colors) positioned relative to the target camera (denote by black). 
On the bottom, from left to right, we show: ground-truth image, predicted novel image, and predicted novel depth, all from the target viewpoint. 
We emphasize that novel images and depth maps are generated \emph{directly as an output of our diffusion model}, rather than rendered from a 3D neural field or set of 3D Gaussians. 

To highlight the multi-view consistency of \Acronym, in Figure \ref{fig:surround} we show qualitative results obtained using the same conditioning views to generate multiple predictions from novel viewpoints, and stacking the predicted colored pointclouds together without any post-processing. Each prediction is generated independently, by setting the novel viewpoint as the origin and positioning the conditioning views relative to it. Even so, they yielded highly consistent pointclouds, both in terms of appearance as well as reconstructed 3D geometry. We attribute this consistency (and ablate it in Table 5 of the main paper) to our proposed scene scale normalization (SSN) procedure, that promotes the generation of depth maps that share the same scale as the one provided by conditioning cameras, even in very different settings (e.g., driving, indoors, object-centric). 

\section{Implicit Dynamics Modeling}
\label{sec:dynamics}

Although \Acronym does not explicitly model dynamic objects, we elected to include datasets with such behavior to increase the diversity of our training data, and report non-trivial improvements relative to a baseline that only considers static datasets (Table 8 of the main paper). 
We attribute this behavior to a learned robustness to the presence of dynamic objects~\cite{zhang2024monst3r}, similar to other methods that rely on self-supervised multi-view consistency with a static environment assumption~\cite{monodepth2,packnet, shu2020featdepth}. 

However, upon further inspection we observed some degree of implicit motion understanding in our learned representation.
Examples are shown in Figure \ref{fig:motion}, using the DDAD~\cite{packnet} dataset. 
In those examples, every $10$th frame from a $100$-frame sequence was used as conditioning, and remaining cameras were used to generate novel images and depth maps. 
As we can observe, moving cars are correctly rendered in different locations to ensure a smooth transition between frames, while static portions of the environment are rendered in the same location, taking into consideration only camera motion. %

\section{Incremental Conditioning}
\label{sec:incremental}

Here we explore how \Acronym scales in terms of the number of conditioning views. Due to the use of latent tokens, computational complexity is largely independent of the number of input tokens, which enables (a) pixel-level diffusion without the need for dedicated auto-encoders; and (b) the simultaneous use of more conditioning views. In fact, one target $256 \times 256$ image generates $65536$ prediction tokens, while each conditioning views adds only $4096$ scene tokens, since image features are produced at $1/4$ the original resolution. In contrast, our largest model has $2048$ latent tokens, which is only $3\%$ of the number of prediction tokens.

In Figure \ref{fig:interpolation} we show the impact of using more conditioning views over a $1267$-frame ScanNet sequence, in terms of novel view (PSNR) and depth (AbsRel) synthesis. We take every $N$-th frame as conditioning views (given by the legend number), and generate predictions for all remaining frames, using the same model from all experiments reported in the main paper. As expected, results degrade in areas further away from available views, and consistently improve as more conditioning views are provided, eventually plateauing at around $100$ (stride $20$). Interestingly, independent experiments using subsets of the sequence ($5$ subsets of $250$ frames) yielded worse results, as evidence that large-scale conditioning (i) does not degrade local performance; and (ii) provides better global context for local predictions.  

As mentioned in Section 3.6 of the main paper, we also take advantage of this highly efficient architecture to investigate how images generated from novel viewpoints can be added as additional conditioning views, thus increasing the amount of available information for future generations. This incremental conditioning strategy should further improve multi-view consistency in cases where model stochasticity becomes relevant, since each novel view is generated independently and thus might come from different parts of the underlying distribution, especially in unobserved areas.
Figure \ref{fig:incremental} provides a quantitative evaluation of our proposed incremental conditioning strategy in terms of novel view (PSNR) and depth (AbsRel) synthesis, compared to the use of a fixed number of conditioning views. As we can observe, the introduction of additional conditioning from generated views consistently improves generation quality, 

In Figure \ref{fig:supp_qualitative_all} we qualitatively show incremental conditioning results on different sequences. Red cameras serve as initial conditioning, and novel images and depth maps are generated from green cameras. After each generation, the predicted image is used as additional conditioning. Since generation order matters in this setting, each new generation is performed on the green camera closest to the initial set of conditioning cameras, that still has not been processed.
Note that \emph{all} previously generated views are used as additional conditioning, which in some scenarios could lead to thousands of images. Even so, we were able to generate novel predictions on a single A100 GPU with 40GB. In terms of inference speed, generations with 25 conditioning views in this setting take $0.5$s, and generations with $1250$ ($50\times$) conditioning views take around $20$s ($40\times$). Additional heuristics, such as using only generated views close to the target view, should lead to increased efficiency while still improving generation quality.

\section{Limitations}

A limitation of our proposed Scene Scale Normalization (SSN) procedure is its inability to simultaneously generate predictions from multiple viewpoints, since the target camera is always assumed to be at the origin. In Section \ref{sec:incremental} we describe an incremental conditioning strategy that mitigates stochasticity when generating predictions from unobserved regions, leading to multi-view consistency over very long sequences ($2000+$ frames). Another current limitation of \Acronym is the lack of dynamics modeling. In Section \ref{sec:dynamics} we show some evidence of implicit modeling of moving objects, however the proper handling of dynamic scenes (e.g., via temporal embeddings and motion tokens, such as ~\cite{anonymous2024storm}) could lead to improvements and spatio-temporal control over novel view and depth synthesis. Moreover, we believe the lack of large-scale dynamic datasets with accurate camera information still constitutes a challenge for the generation of such spatio-temporal implicit foundation model.

{
\small
\bibliographystyle{ieeenat_fullname}
\bibliography{references}
}

\end{document}